\documentclass[conf]{IEEEtran}
\usepackage{algorithm,algpseudocode}
\usepackage{amsmath}
\usepackage{amssymb}
\usepackage{mathtools}
\usepackage[dvipsnames]{xcolor}
\usepackage{amssymb}
\usepackage{graphicx}
\usepackage{amsmath}
\usepackage{amssymb}
\usepackage{booktabs}

\usepackage{changepage}
\usepackage{extramarks}
\usepackage{fancyhdr}
\usepackage{lastpage}
\usepackage{setspace}
\usepackage{soul}
\usepackage{xspace}
\usepackage{amssymb}

\usepackage{adjustbox}
\usepackage{array}
\usepackage{booktabs}
\usepackage{colortbl}
\usepackage{float,wrapfig}
\usepackage{hhline}
\usepackage{multirow}
\usepackage{subcaption} 
\usepackage{mathtools}

\usepackage{algorithm, algpseudocode}
\usepackage{enumerate}
\usepackage{lipsum}

\usepackage{listings}

\usepackage{enumitem}
\usepackage{graphics}
\newcommand{\etal}{\textit{et al}. }
\newcommand{\ie}{\textit{i}.\textit{e}., }
\newcommand{\eg}{\textit{e}.\textit{g}., } 

\usepackage{pifont}
\usepackage{cite}
\def\BibTeX{{\rm B\kern-.05em{\sc i\kern-.025em b}\kern-.08em
    T\kern-.1667em\lower.7ex\hbox{E}\kern-.125emX}}
\usepackage[bookmarksnumbered=true]{hyperref} 
\hypersetup{
     colorlinks = true,
     linkcolor = blue,
     anchorcolor = blue,
     citecolor = blue,
     filecolor = blue,
     urlcolor = blue
     }

\usepackage{graphicx}
\usepackage{svg}

\usepackage{wrapfig}
\usepackage{caption}
\usepackage{subcaption}
\usepackage{booktabs}

\usepackage{tikz}

\usepackage{ragged2e}
\usepackage{siunitx}
\usepackage{multicol}
\usepackage{multirow}

\usepackage{xcolor}
\begin{document}

\title{Sharing Leaky-Integrate-and-Fire Neurons for Memory-Efficient Spiking Neural Networks}

\author{ Youngeun Kim, \textit{Student Member, IEEE}, Yuhang Li, \textit{Student Member, IEEE}, Abhishek Moitra, \textit{Student Member, IEEE}, Ruokai Yin, \textit{Student Member, IEEE} and Priyadarshini Panda, \textit{Member, IEEE}
\thanks{ Youngeun Kim, Yuhang Li, Abhishek Moitra, Ruokai Yin and Priyadarshini Panda are with
the Department of Electrical Engineering, Yale University, New Haven, CT,
USA. }
}

\maketitle

\begin{abstract}
  Spiking Neural Networks (SNNs) have gained increasing attention as energy-efficient neural networks owing to their binary and asynchronous computation. However, their non-linear activation, that is Leaky-Integrate-and-Fire (LIF) neuron, requires additional memory to store a membrane voltage to capture the temporal dynamics of spikes. Although the required memory cost for LIF neurons significantly increases as the input dimension goes larger, a technique to reduce memory for LIF neurons has not been explored so far. To address this, we propose a simple and effective solution, EfficientLIF-Net, which shares the LIF neurons across different layers and channels. Our EfficientLIF-Net achieves comparable accuracy with the standard SNNs while bringing up to $\sim 4.3\times$ forward memory efficiency and $\sim 21.9\times$ backward memory efficiency for LIF neurons. We conduct experiments on various datasets including CIFAR10, CIFAR100, TinyImageNet, ImageNet-100, and N-Caltech101. Furthermore, we show that our approach also offers advantages on Human Activity Recognition (HAR) datasets, which heavily rely on temporal information.
\end{abstract}




\maketitle

\section{Introduction}

Spiking Neural Networks (SNNs) have gained significant attention as a promising candidate for low-power machine intelligence \cite{roy2019towards,christensen20222022,wu2018spatio,wu2019direct,kundu2021hire,fang2021deep}.
By mimicking biological neuronal mechanisms, Leaky-Integrate-and-Fire (LIF) neurons in SNNs convey visual information with temporal binary spikes over time.
The LIF neuron \cite{liu2001spike} considers temporal dynamics by accumulating incoming spikes inside a membrane potential, and generates output spikes when the membrane potential voltage exceeds a firing threshold. 
Such binary and asynchronous operation of SNNs incurs energy-efficiency benefits on low-power neuromorphic hardware \cite{akopyan2015truenorth,davies2018loihi,furber2014spinnaker,orchard2021efficient}.

\begin{figure}[t]
\begin{center}
\def\arraystretch{0.5}
\begin{tabular}{@{}c@{\hskip 0.03\linewidth}c@{}c}
\includegraphics[width=0.86\linewidth]{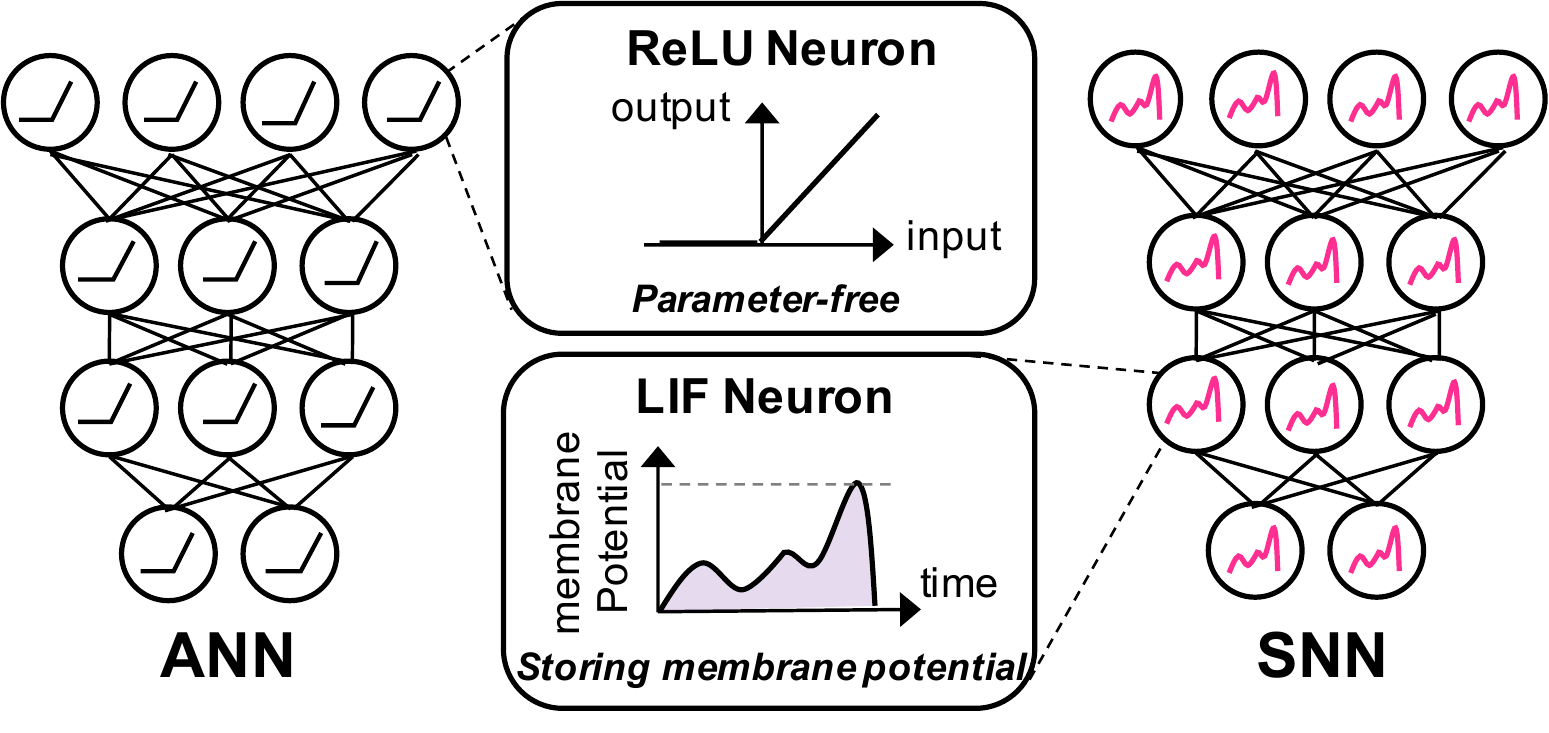} \\
\includegraphics[width=0.75\linewidth]{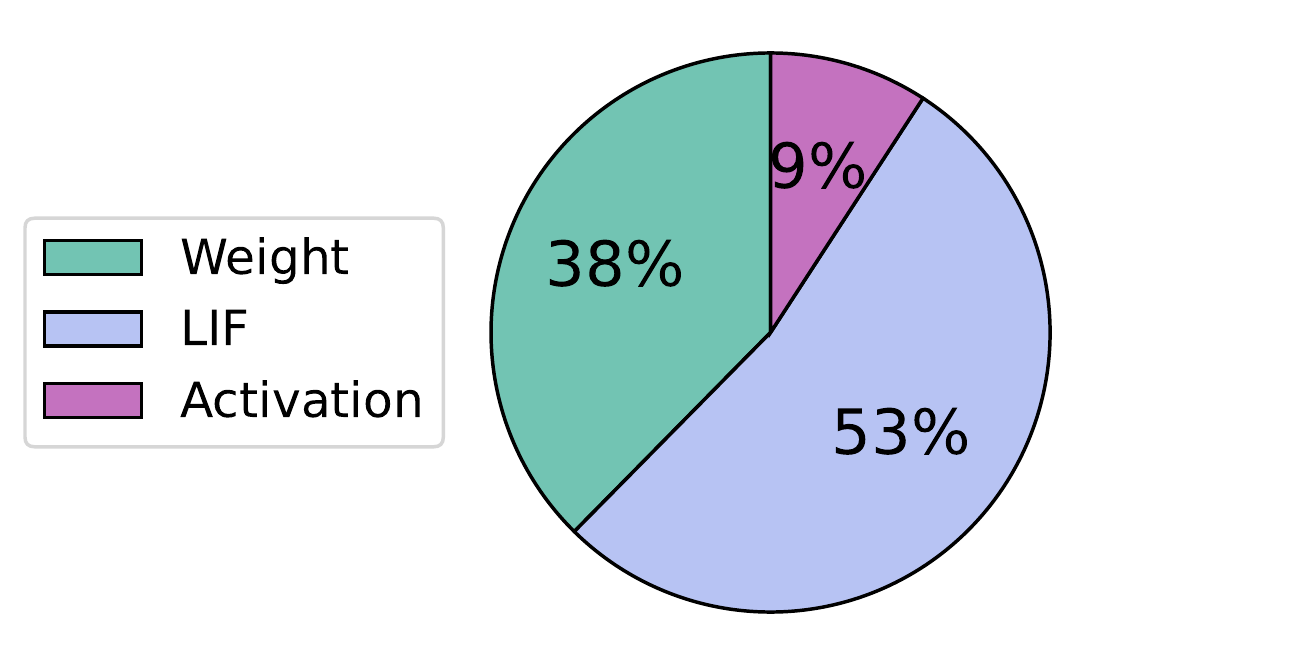} 
\end{tabular}
\end{center}
\caption{
Motivation of our work. 
\textbf{Top:} Comparison between neurons in ANNs and SNNs: Unlike ReLU neurons, which do not require any parameters, LIF neurons maintain a membrane potential with voltage values that change across timesteps.
\textbf{Bottom:} Memory cost breakdown for the Spiking-ResNet19 architecture during inference on an image with a resolution of $224\times 224$.
}
\label{fig:intro:lif_story}
\end{figure}

Although SNN brings computational efficiency benefits, memory overhead caused by LIF neurons can be problematic. 
As shown in Fig. \ref{fig:intro:lif_story}, LIF neurons require additional memory for storing the membrane potential value which changes over time. 
This is not the case for the traditional Artificial Neural Networks (ANNs) where most non-linear activation functions are parameter-free (\eg ReLU, Sigmoid).
At the same time, LIF neurons occupy a large portion of memory with the high-resolution input image (Fig. \ref{fig:intro:lif_story}). 
For instance, the LIF memory takes $53\%$ of memory overhead in the case of ResNet19 \cite{he2016deep} with a $224 \times 224$ image.
Unfortunately, the LIF memory overhead has been overlooked so far in SNN studies.

To address this, we propose EfficientLIF-Net where we share the LIF neurons across different layers and channels.
By sharing the memory, we do not need to assign separate memory for each layer and channel.
For layer-wise sharing, we use common LIF neurons across layers having the same activation size, such as layers in one ResNet block \cite{he2016deep}.
For channel-wise sharing, we equally divide the LIF neurons into multiple groups through the channel dimension and share common LIF neurons across different groups.
Surprisingly, our EfficientLIF-Net provides similar performance as the standard SNN models where each layer and channel has independent LIF neurons.
We show the gradient can successfully flow back through all layers, thus the weight can be trained to consider the temporal dynamics of spike information.

Furthermore, the proposed EfficientLIF-Net brings huge benefits to saving memory costs during training. 
Spatio-temporal operation inside SNNs incurs a huge computational graph for computing backward gradients.
Each LIF neuron needs to store membrane potential to make gradients flow back, where the training memory increases as the SNN goes deeper and uses larger timesteps.
This huge computational graph often is difficult to be trained on the limited GPU memory \cite{liang2021h2learn,singh2022skipper,yin2022sata}.
In this context, since our architecture shares the membrane potential across all layers, we can compute each layer’s membrane potential from the next layer’s membrane potential real-time during backward step.
This enables us to perform backpropagation without the need for storing/caching the membrane potentials of all layers in memory (from the forward step).

Our contributions can be summarized as follows: 
\begin{itemize}
\item We pose the memory overhead problem of LIF neurons in SNNs, where the memory cost significantly increases as the image size goes larger.
\item To address this, we propose a simple and effective architecture, EfficientLIF-Net where we share the LIF neurons across different layers and channels. 
%
\item EfficientLIF-Net also reduces memory cost during training by computing each layer’s (channel's) membrane potential from the next layer’s (channel's) membrane potential real-time during backward step, drastically reducing the caching of membrane potentials.
\item We conduct experiments on five public datasets, validating EfficientLIF-Net can achieve comparable performance as the standard SNNs while bringing up to $\sim 4.3\times$ forward memory efficiency and up to $\sim 21.9\times$ backward memory efficiency for LIF neurons.
\item We also observe that the LIF memory cost problem exists in pruned SNNs and in fact the LIF memory overhead percentage goes higher when the weight sparsity goes higher. 
Our EfficientLIF-Net successfully reduces the LIF memory cost to $\sim  23\%$ in pruned SNNs while achieving iso-accuracy compared to the pruned baseline.
%
\end{itemize}

\section{Related Work}
\subsection{Spiking Neural Networks}
\vspace{-1mm}
Different from the standard Artificial Neural Networks (ANNs), Spiking Neural Networks (SNNs) convey temporal spikes \cite{roy2019towards,christensen20222022}.
Here, Leaky-Integrate-and-Fire (LIF) neuron plays an important role as the non-linear activation.
The LIF neurons have a ``memory" called membrane potential, where the incoming spikes are accumulated.
Output spikes are generated if the membrane potential exceeds a firing threshold, then the membrane potential resets to zero.
This firing operation of LIF neurons is non-differentiable, so the previous SNN literature has focused on resolving the gradient problem.
A widely-used training technique is converting pre-trained ANNs to SNNs using weight or threshold balancing \cite{sengupta2019going,han2020rmp,diehl2015fast,rueckauer2017conversion,li2021free}. However, such methods require large number of timesteps to emulate float activation using binary spikes.
Recently, a line of works propose to circumvent the non-differentiable backpropagation problem by defining a surrogate function  \cite{lee2016training,lee2020enabling,neftci2019surrogate,shrestha2018slayer,wu2018spatio,wu2021training,li2021differentiable,kim2022neural,wu2020progressive,kim2021revisiting,kim2022exploringb}.
As the weight is trained to consider temporal dynamics, they show both high performance and short latency. 
Although the previous methods have made huge advances in terms of improving the performance, they assume that SNNs have different LIF neurons for different layers and channels, which imposes a huge memory overhead in both forward and backward.

\subsection{Compression Methods for Efficient SNNs}
\vspace{-1mm}
Due to the energy-efficiency benefit of SNNs, they can be  suitably implemented on edge devices with limited memory storage \cite{venkatesha2021federated,yang2022lead,skatchkovsky2020federated}.
Therefore, a line of work has proposed various methods to reduce the memory cost for SNNs using compression techniques.
Neural pruning is one of the effective methods for SNN compression.
Several works \cite{neftci2016stochastic,rathi2018stdp} have proposed a post-training pruning technique using a threshold value.
Unsupervised online adaptive weight pruning \cite{guo2020unsupervised}  dynamically prunes trivial weights over time. 
Shi \etal \cite{shi2019soft} prune weight connections during training with a soft mask.
Recently, deeper SNNs are pruned with ADMM optimization tool \cite{deng2021comprehensive}, gradient-based rewiring \cite{chen2021pruning}, and lottery ticket hypothesis \cite{kim2022exploring}.
Meanwhile, various quantization techniques also have been proposed to compress SNNs \cite{li2022quantization,meng2022training,guo2022reducing,datta2022hoyer}.
Schaefer and Joshi  \cite{schaefer2020quantizing} propose integer fixed-point representations for neural dynamics, weights, loss, and gradients.
The recent work \cite{chowdhury2021spatio} performs quantization through temporal dimension for low-latency SNNs.
Lui and Neftci propose a quantization technique based on the Hessian of weights \cite{lui2021hessian}.
Nonetheless, no prior work has explicitly addressed the memory overhead caused by LIF neurons.
Our method effectively reduces memory overhead by modifying the architecture, and is orthogonal to previous methods. Thus, combining EfficientLIF-Net with compression techniques will further compound the benefits.

\begin{figure*}[t]
\begin{center}
\def\arraystretch{0.5}
\begin{tabular}{@{\hskip 0.003\linewidth}c@{\hskip 0.003\linewidth}c@{\hskip 0.05\linewidth}c@{\hskip 0.01\linewidth}c@{}c}

\includegraphics[width=0.118\linewidth]{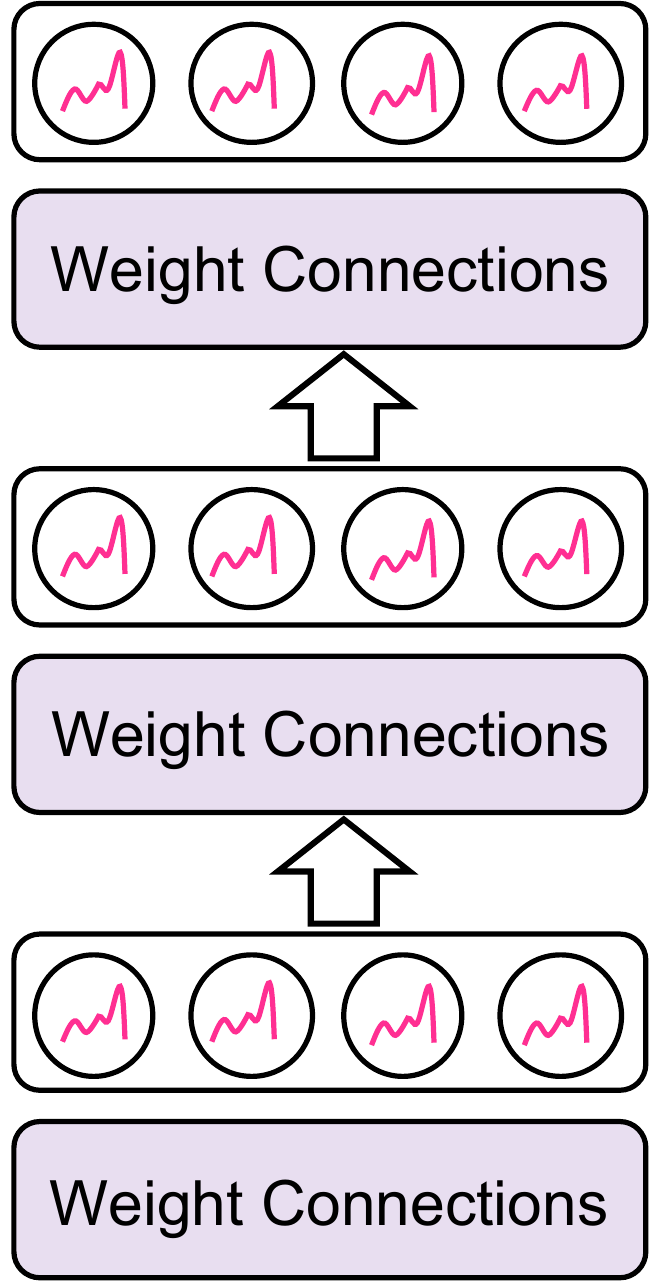} &
\includegraphics[width=0.161\linewidth]{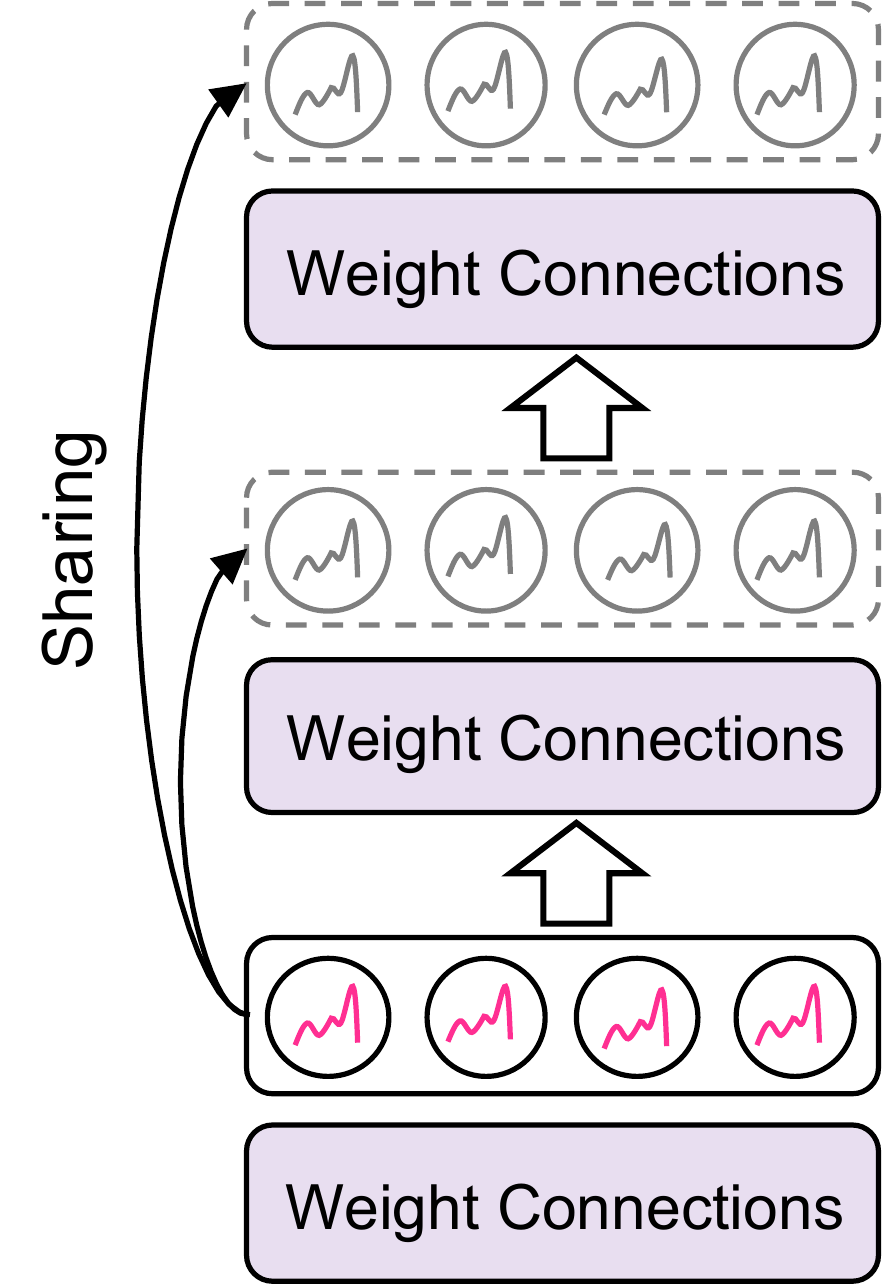} &
\includegraphics[width=0.200\linewidth]{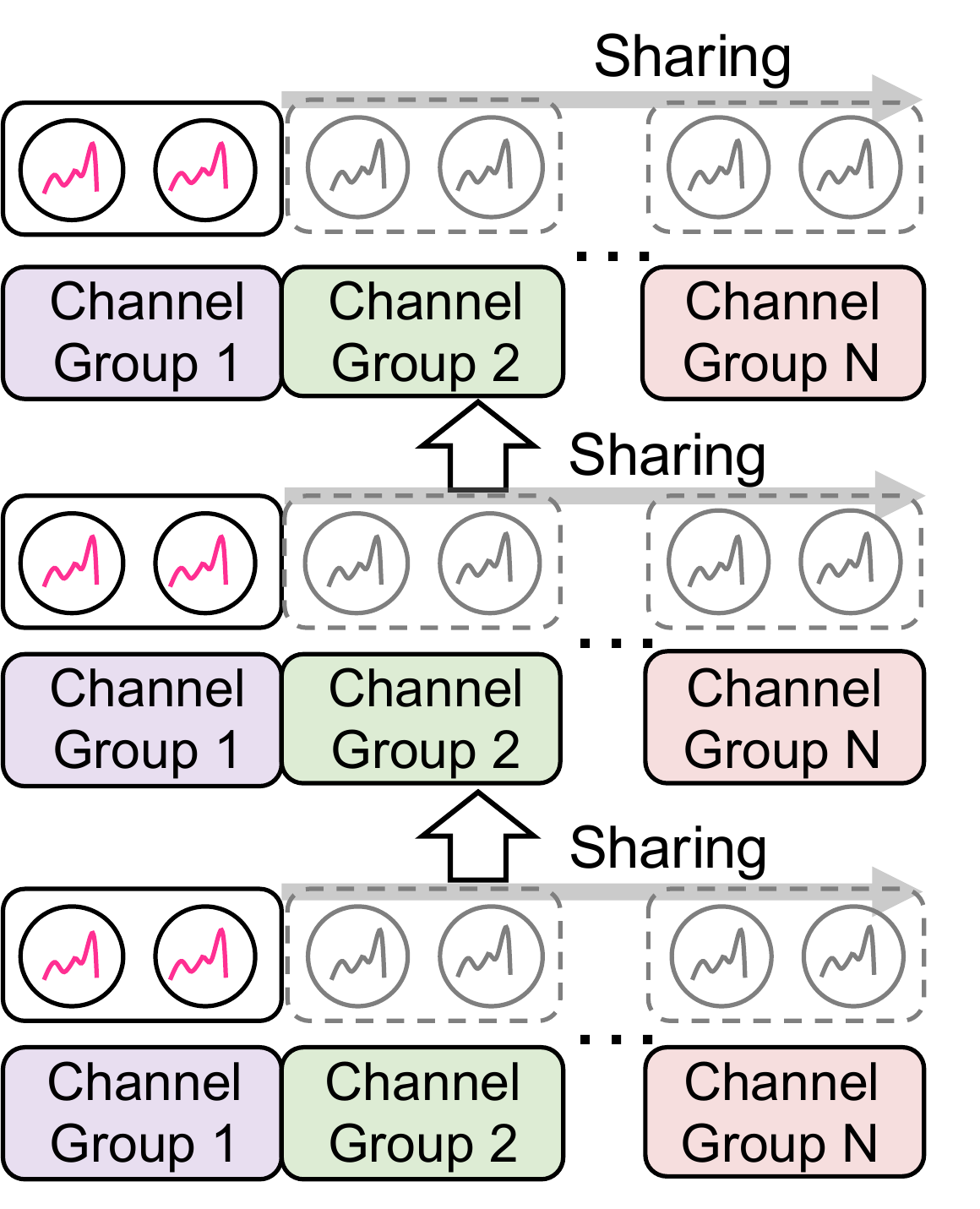} &
\includegraphics[width=0.249\linewidth]{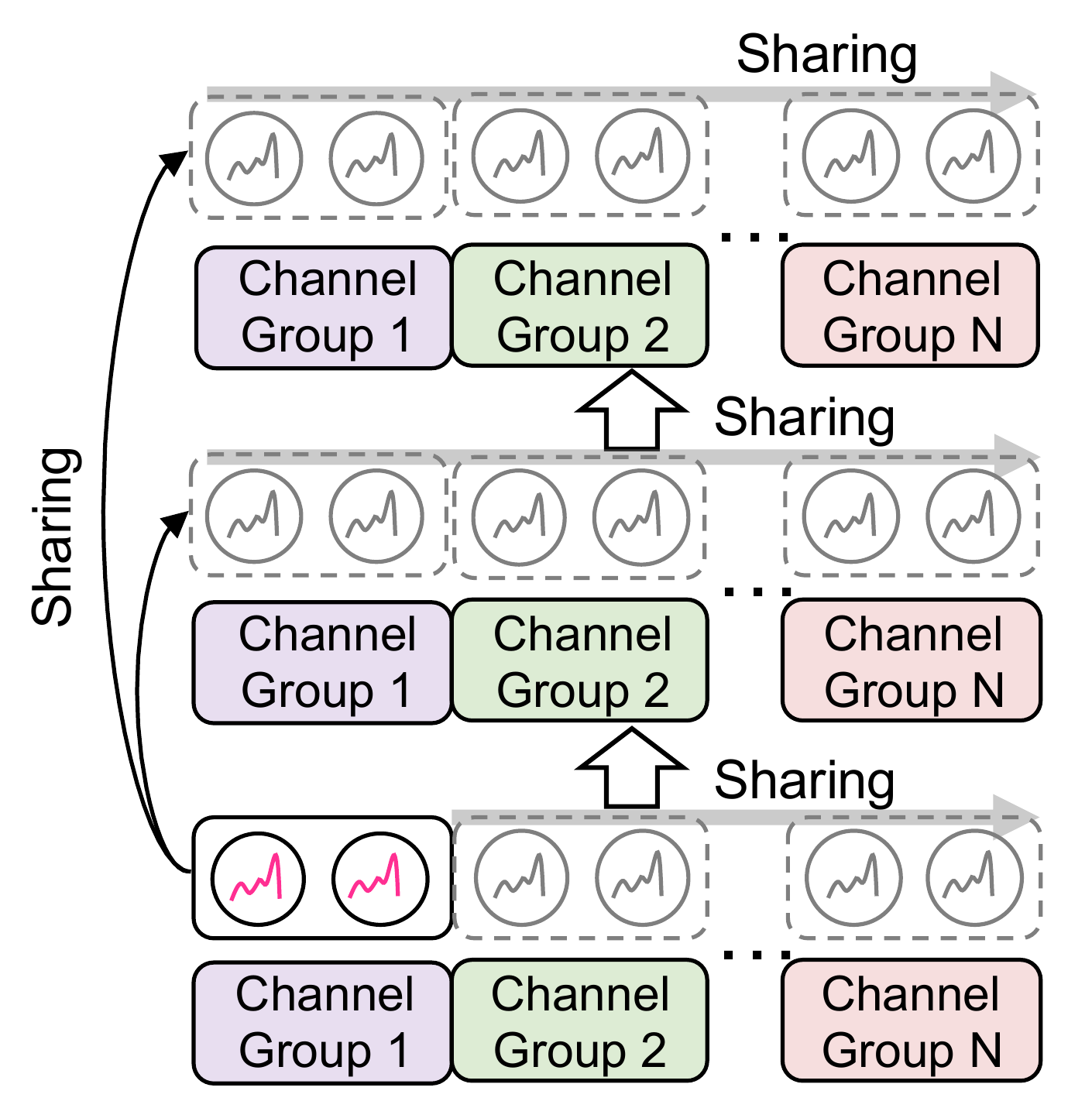} 
\\
\\
 {(a) Baseline SNN } & \hspace{7mm}
{(b) Cross-layer Sharing } & \hspace{-2mm}{(c) Cross-channel Sharing}  & \hspace{7mm}{(d) Cross-layer\&channel Sharing} \\
\end{tabular}
\caption{ 
Illustration of the proposed EfficientLIF-Net. (a) Conventional SNNs where each layer and channel has separate LIF neurons. (b)-(d) is our proposed EfficientLIF-Net which shares LIF neurons across layer, channel, and layer \& channel.}
\label{fig:method:forward}
\end{center}
\end{figure*}

\section{Preliminaries}

\subsection{Leaky Integrate-and-Fire Neuron}

In our paper, we mainly address the memory cost from a Leaky-Integrate-and-Fire (LIF) neuron, which is widely adopted in SNN works \cite{lee2020enabling,wu2018spatio,wu2021training,li2021differentiable,kim2022neural,wu2020progressive,li2021free,fang2021deep,fang2021incorporating}.
Suppose LIF neurons in $l$-th layer have membrane potential $U_l^t$ at timestep $t$, we can formulate LIF neuron dynamics as:
\begin{equation}
    U_l^t = \lambda  U_l^{t-1} + W_{l}O_{l-1}^t, 
    \label{eq:LIF}
\end{equation}
where $W_{l}$ is weight parameters in layer $l$, $O_{l-1}^t$ represents the spikes from the previous layer, $\lambda$ is a decaying factor in the membrane potential.
Note, we use uppercase letters for matrix notation.
The LIF neuron generates an output spike $O_l^{t}$ when the membrane potential exceeds the firing threshold $\theta$. Here, we define the spike firing function as:
\begin{equation}
f(U_l^t) = O_{l}^t = 
    \begin{cases}
        1 & \text{if  } U_l^t>\theta\\
        0 & \text{otherwise}
    \end{cases}.
\label{eq:fire}
\end{equation}
After firing, the membrane potential can be reset to zero (\ie hard reset), or reduced by the threshold value (\ie soft reset).
Thus, an LIF neuron always stores the membrane potential to capture the temporal information of spikes.
The memory cost for LIF neurons is proportional to the input image dimension, which poses huge memory overhead for high-resolution data such as ImageNet \cite{deng2009imagenet}.

\subsection{Gradient Backpropagation in SNNs}

\begin{figure*}[t]
\begin{center}
\def\arraystretch{0.5}
\begin{tabular}{@{\hskip 0.0\linewidth}c@{\hskip 0.015\linewidth}c@{\hskip 0.015\linewidth}c@{\hskip 0.015\linewidth}c@{}c}

\includegraphics[width=0.22\linewidth]{ 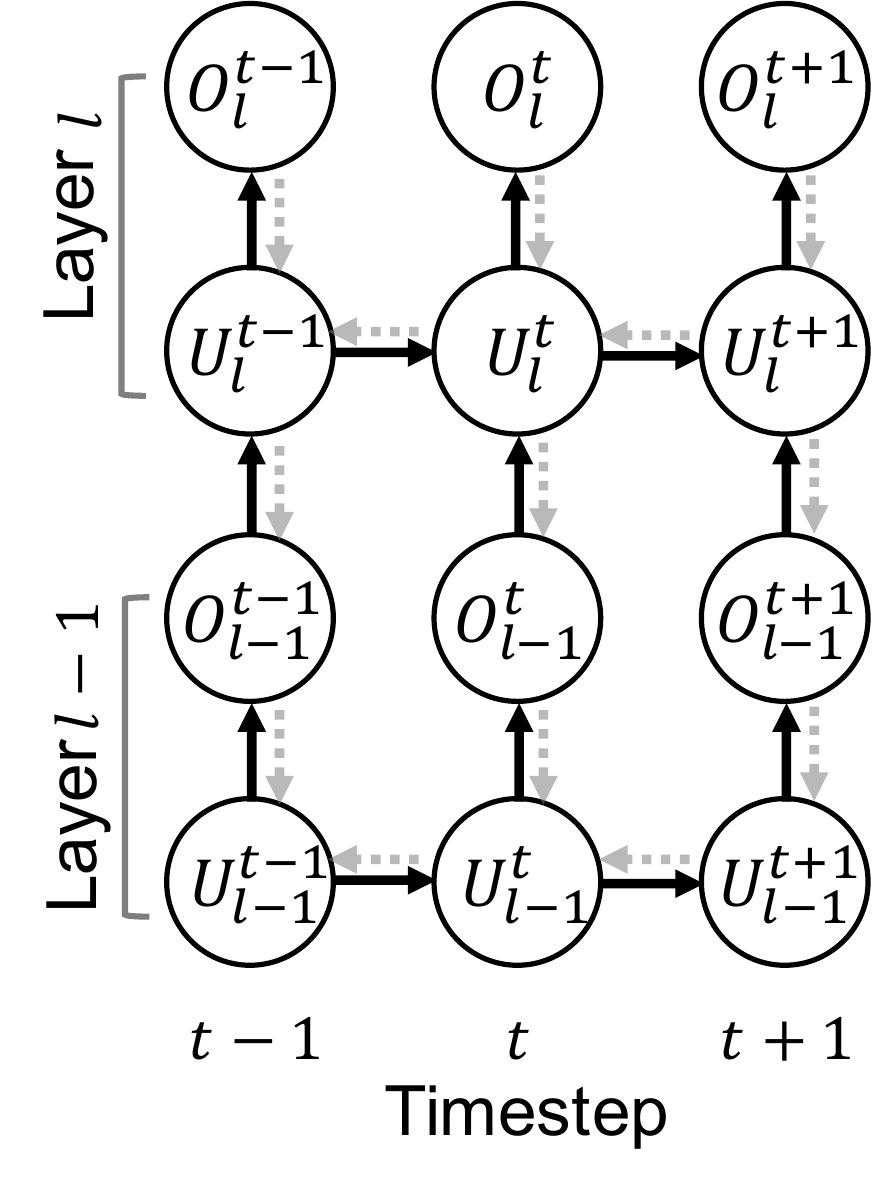} &
\includegraphics[width=0.23\linewidth]{ 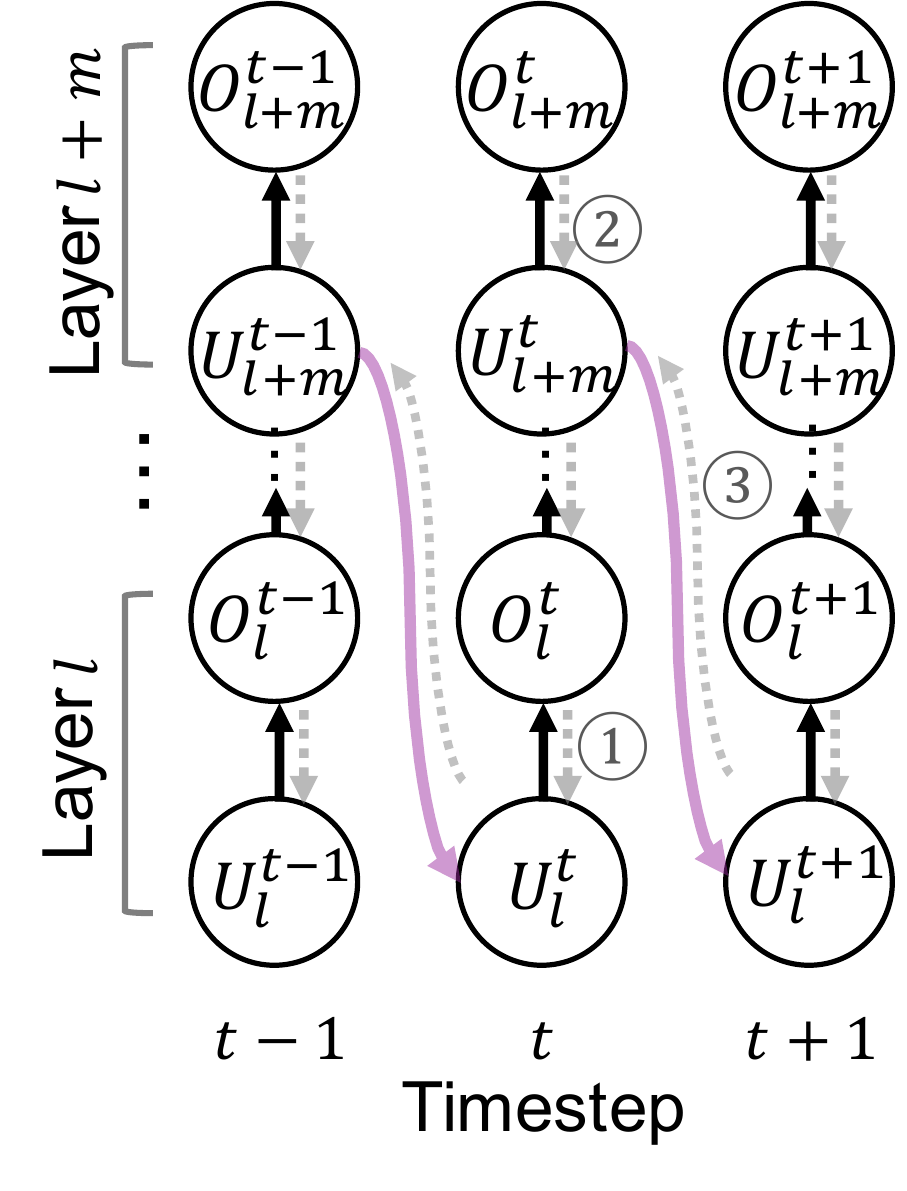} &
\includegraphics[width=0.46\linewidth]{ 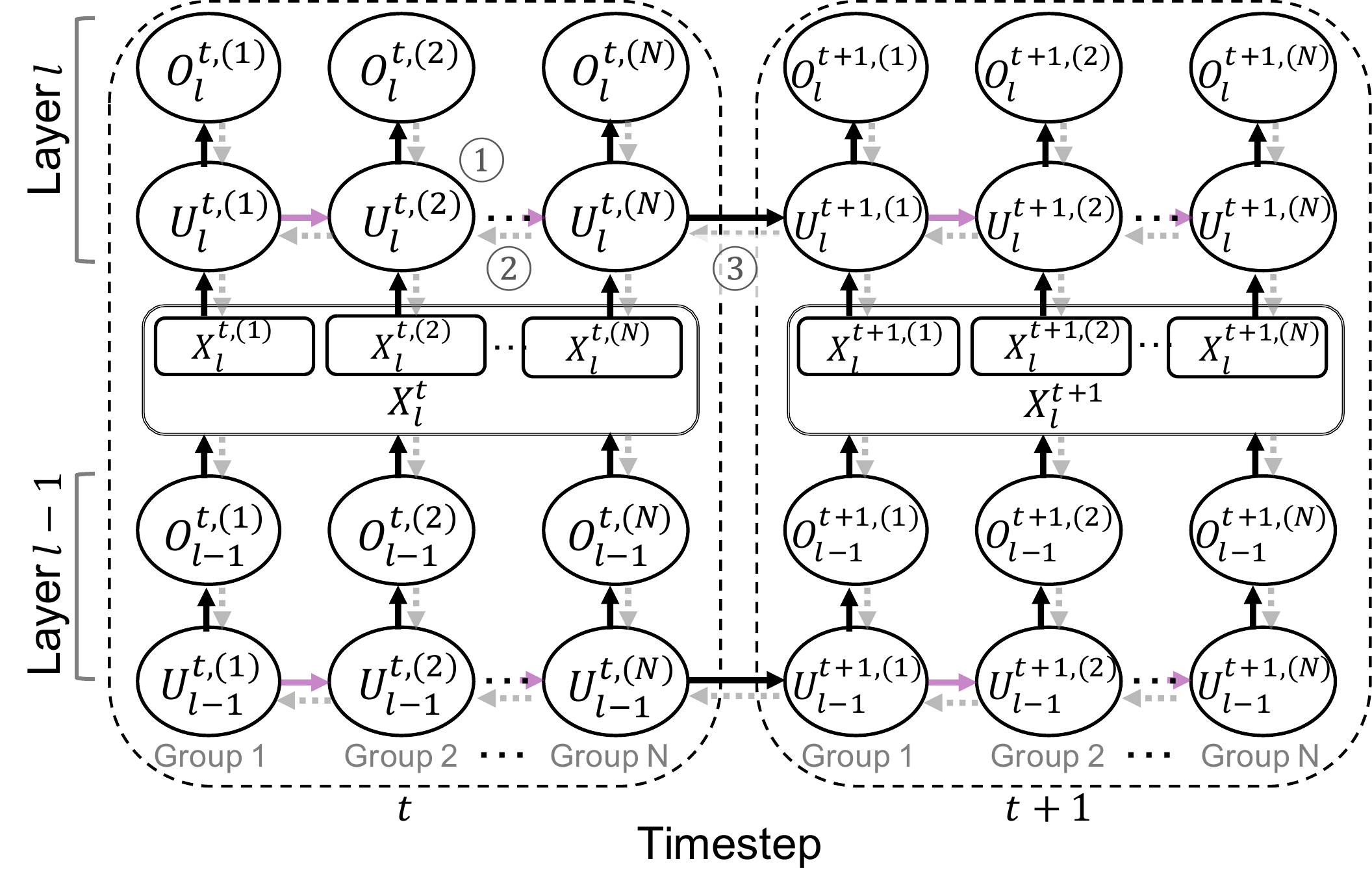} 
\\
\hspace{4mm} {(a) Baseline SNN } & \hspace{7mm}
{(b) Cross-layer Sharing } & \hspace{4mm}{(c) Cross-channel Sharing}  \\
\end{tabular}
\caption{ Illustration of a unrolled computational graph for the backpropagation.  Black solid arrows and gray dotted arrows represent forward and backward paths, respectively.
For simplicity, we omit the reset path from the spike output.
}
 \vspace{-1mm}
\label{fig:method:backward}
\end{center}
\end{figure*}

For the class probability prediction, we accumulate the final-layer activation across all timesteps, followed by the Softmax function.
We apply cross-entropy loss $L$ for training the weights parameters.
The backward gradients are calculated in both spatial and time axis \cite{wu2018spatio,neftci2019surrogate} according to the chain rule:
\begin{equation}
      \frac{\partial L}{\partial W_l} = \sum_{t}(\frac{\partial L}{\partial O_l^t}\frac{\partial O_l^t}{\partial U_l^t} + \frac{\partial L}{\partial U_l^{t+1}}  \frac{\partial U_l^{t+1}}{\partial U_l^{t}})
 \frac{\partial U_l^t}{\partial W_l}.
\label{eq:delta_W}
\end{equation}
%
%
Here, the gradient of output spikes with respect to the membrane potential 
$\frac{\partial O_l^t}{\partial U_l^t}$ is non-differentiable.
Following previous work \cite{fang2021deep}, we use $arctan()$ to approximate gradients, \ie we use an approximate function
$f(x) = \frac{1}{\pi}arctan(\pi x) + \frac{1}{2}$ for computing gradients of $\frac{\partial O_l^t}{\partial U_l^t}$.
The overall computational graph is illustrated in Fig. \ref{fig:method:backward}(a).


\section{EfficientLIF-Net}
In this section, we first describe the details of how we reduce the memory cost of LIF neurons across layers and channels. The overall concept of EfficientLIF-Net is illustrated in Fig. \ref{fig:method:forward}.
After that, we provide the analysis of the backward gradient in EfficientLIF-Net for training, which shows our EfficientLIF-Net successfully considers the entire time horizon.
Finally, we show the memory advantage of our EfficientLIF-Net during backpropagation.

\subsection{Sharing Memory of LIF neurons}

\noindent\textbf{Cross-layer Sharing.} The key idea here is sharing the LIF neurons across different layers where they have the same output activation size.
Thus, LIF neurons are shared across multiple subsequent layers before the layer increases channel size or reduces spatial resolution.
Such architecture design can be easily observed in CNN architectures such as ResNet \cite{he2016deep}.

Let's assume the networks have the same activation size from the $l$-th layer to the $(l+m)$-th layer.
The membrane potential of the $(l+1)$-th layer is calculated by adding the previous layer's membrane potential and weighted spike output from the previous layer: 
\begin{equation}
    U_{l+1}^t = \lambda ({U}_{l}^{t}-{O}_{l}^{t}) + W_{l+1}O_{l}^t.
    \label{eq:LIF_crosslayer}
\end{equation}
Here the previous layer's membrane potential ${U}_{l}^{t}$ decreases its value by the threshold for soft reset (firing threshold is set to $1$) after it generates spikes ${O}_{l}^{t}$.
After that, decay factor $\lambda$ is applied to the previous layer's membrane potential, since we aim to dilute the previous layers' information as networks go deeper.
The layer $(l+1)$ generates output spike following Eq. \ref{eq:fire}:
\begin{equation}
    O_{l+1}^t = f(U_{l+1}^t).
    \label{eq:LIF_crosslayer_fire}
\end{equation}
In the same timestep, the spike information goes through all layers (from $l$-th layer to $l+m$-th layer) with Eq. \ref{eq:LIF_crosslayer} and Eq. \ref{eq:LIF_crosslayer_fire} dynamics. Then, the membrane potential of layer $l+m$ is shared with layer $l$ at the next timestep (purple arrow in Fig. \ref{fig:method:backward}(b)).
\begin{equation}
    U_{l}^{t+1} = \lambda  (U_{l+m}^{t}- {O}_{l+m}^{t}) + W_{l}O_{l-1}^{t+1},  
    \label{eq:LIF_crosslayer_last}
\end{equation}
where the soft reset and decaying is applied to $U_{l+m}^{t}$, and the weighted input comes from layer $l-1$.

Overall, we require only one-layer LIF memory for layer $l \sim$ layer $(l+m)$ computation, which is shared across all layers and timesteps. 
Thus, LIF memory of layers $l \sim (l+m)$ can be reduced by $\frac{1}{m}$.
The overall computational graph is illustrated in Fig. \ref{fig:method:backward}(b).

\noindent\textbf{Cross-channel Sharing.}
We also explore the neuron sharing scheme in the channel dimension.
Let $X_{l}$ be the weighted input spike, \ie $X_{l}=W_{l}O_{l-1}^t$, then we first divide the weighted input spike tensor into $N$ groups in channel axis.
\begin{equation}
 X_{l}^t \rightarrow [X_{l}^{t, (1)}, X_{l}^{t,(2)},  ..., X_{l}^{t,(N)}].
    \label{eq:LIF_crosslchanne_decompose}
\end{equation}
Suppose  $ X_{l}^t \in \mathbb{R}^{C \times H \times W}$, then the spike of each group can be represented as $ X_{l}^{t, (i)} \in \mathbb{R}^{\frac{C}{N} \times H \times W}$, $ i \in \{1, 2, ..., N\}$.
Then, the LIF neurons can be sequentially shared across different groups (\ie different channels) of weighted input spike. The membrane potential of $(i+1)$-th group at layer $l$ can be formulated as:
\begin{equation}
    U_l^{t,(i+1)} = \lambda  (U_l^{t, (i)}-O_{l}^{t, (i)}) + X_{l}^{t, (i+1)},  
    \label{eq:LIF_crosschannel}
\end{equation}
where $U_l^{t, (i)}$ is the membrane potential of the previous group, and $X_{l}^{t, (i+1)}$ is the incoming weighted spike input of the $(i+1)$-th group from the previous layer.
Here, soft reset and decaying also applied.
The output spikes of each group are generated by standard firing dynamics (Eq. \ref{eq:fire}):
\begin{equation}
    O_{l}^{t,(i)} = f(U_l^{t,(i)}).
    \label{eq:LIF_crosschannel_fire}
\end{equation}
We concatenate the output spikes of each groups through channels in order to compute the output at timestep $t$:
\begin{equation}
 O_{l}^t = [O_{l}^{t, (1)}, O_{l}^{t,(2)},  ..., O_{l}^{t,(N)}].
    \label{eq:LIF_crosschannel_outconcat}
\end{equation}
After completing the LIF sharing in timestep $t$,  we share the last group's (\ie group $N$) membrane potential to the first group in the next timestep $t+1$.
\begin{equation}
    U_l^{t+1,(1)} = \lambda  (U_l^{t, (N)}-O_{l}^{t, (N)}) + X_{l}^{t+1, (1)}. 
    \label{eq:LIF_crosschannel_nexttime}
\end{equation}

By using cross-channel sharing, the memory cost for LIF neuron of one layer can be reduced by $\frac{1}{N}$, where $N$ is the number of groups. Thus, memory-efficiency will increase as we use larger group number.

\noindent\textbf{Cross-layer\&channel Sharing.}
The cross-layer and cross-channel sharing methods are complementary to each other, therefore they can be used together to bring further memory efficiency.
The LIF neurons are shared across channels and layers as shown in Fig. \ref{fig:method:forward}(d). The neuron-sharing mechanism can be obtained by combining cross-layer and cross-channel sharing methods. 

Let's assume the networks have the same activation size from the $l$-th layer to the $(l+m)$-th layer.
The sharing mechanism in one layer is same as channel sharing.
Let $X_{l}$ be the weighted input spike, \ie $X_{l}=W_{l}O_{l-1}^t$, then we first divide the weighted input spike tensor into $N$ groups in channel axis.
\begin{equation}
 X_{l}^t \rightarrow [X_{l}^{t, (1)}, X_{l}^{t,(2)},  ..., X_{l}^{t,(N)}].
    \label{eq:LIF_crosslchanne_decompose}
\end{equation}
Suppose  $ X_{l}^t \in \mathbb{R}^{C \times H \times W}$, then the spike of each group can be represented as $ X_{l}^{t, (i)} \in \mathbb{R}^{\frac{C}{N} \times H \times W}$, $ i \in \{1, 2, ..., N\}$.
Then, the LIF neurons can be sequentially shared across different groups (\ie different channels) of weighted input spike. The membrane potential of $(i+1)$-th group at layer $l$ can be formulated as:
\begin{equation}
    U_l^{t,(i+1)} = \lambda  (U_l^{t, (i)}-O_{l}^{t, (i)}) + X_{l}^{t, (i+1)},  
    \label{eq:LIF_crosschannellayer}
\end{equation}
where $U_l^{t, (i)}$ is the membrane potential of the previous group, and $X_{l}^{t, (i+1)}$ is the incoming weighted spike input of the $(i+1)$-th group from the previous layer.
Here, soft reset and decaying is also applied.
The output spikes of each group are generated by standard firing dynamics:
\begin{equation}
    O_{l}^{t,(i)} = f(U_l^{t,(i)}).
    \label{eq:LIF_crosschannellayer_fire}
\end{equation}
We concatenate the output spikes of each group through channels in order to compute the output at timestep $t$:
\begin{equation}
 O_{l}^t = [O_{l}^{t, (1)}, O_{l}^{t,(2)},  ..., O_{l}^{t,(N)}].
    \label{eq:LIF_crosschannellayer_outconcat}
\end{equation}
After completing the LIF sharing at layer $l$,  we share the last group's (\ie group $N$) membrane potential of $l$-th layer to the first group of $l+1$-th layer.
\begin{equation}
    U_{l+1}^{t,(1)} = \lambda  (U_l^{t, (N)}-O_{l}^{t, (N)}) + X_{l+1}^{t+1, (1)}. 
    \label{eq:LIF_crosschannellayer_nexttime}
\end{equation}
In the same timestep, the spike information goes through all layers (from $l$-th layer to $l+m$-th layer) dynamics. Then, the last group's (\ie group $N$) membrane potential of layer $l+m$ is shared with the first group of layer $l$ at the next timestep.
\begin{equation}
    U_l^{t+1,(1)} = \lambda  (U_{l+m}^{t, (N)}-O_{l+m}^{t, (N)}) + X_{l}^{t+1, (1)}. 
    \label{eq:LIF_crosschannellayer_nexttime}
\end{equation}

By using cross-channel sharing, the memory cost of LIF neuron for layer $l \sim$ layer $(l+m)$ computation can be reduced by $\frac{1}{mN}$, where $N$ is the number of groups.
Our experimental results show that although we combine two sharing methods, we still get iso-accuracy as the standard SNNs.

\subsection{Gradient Analysis}
Sharing LIF neurons leads to different gradient paths compared to standard SNNs. 
Therefore, we provide the gradient analysis for EfficientLIF-Net.

\noindent\textbf{Gradient of Cross-layer Sharing.}
Suppose that we compute the gradients for $m$ subsequent layers where they have the same activation size.
For simplicity, we call these $m$ subsequent layers as a ``sharing block".
The unrolled computational graph is illustrated in Fig. \ref{fig:method:backward}(b).

For the intermediate layers of the sharing block, the gradients flow back from the next layer (marked as \raisebox{.5pt}{\textcircled{\raisebox{-.9pt} {1}}} in Fig. \ref{fig:method:backward}(b)), which can be formulated as:
\begin{equation}
      \frac{\partial L}{\partial W_l} = 
      \sum_{t}
      \left(
      \frac{\partial L}{\partial O_l^t}
      \frac{\partial O_l^t}{\partial U_l^t} 
      +
      \frac{\partial L}{\partial U_{l+1}^{t}} 
      \frac{\partial U_{l+1}^{t}}{\partial U_l^{t}}
      \right)
       \frac{\partial U_l^t}{\partial W_l},
\label{eq:bw_crosslayer_1}
\end{equation}
where both terms are derived by the forward dynamics in Eq. \ref{eq:LIF_crosslayer}.
For the final layer of the sharing block, the gradients flow back through both layer and temporal axis:
\begin{equation}
\hspace{-1mm}
      \frac{\partial L}{\partial W_{l+m}} 
      \hspace{-1mm}
      = \hspace{-1mm}
      \sum_{t}
      \left(
      \frac{\partial L}{\partial O_{l+m}^t}
      \frac{\partial O_{l+m}^t}{\partial U_{l+m}^t}
      + 
      \frac{\partial L}{\partial U_{l}^{t+1}}
      \frac{\partial U_{l}^{t+1}}{\partial U_{l+m}^{t}}\right)
      \frac{\partial U_{l+m}^t}{\partial W_{l+m}}.
\label{eq:bw_crosslayer_2}
\end{equation}
The first term shows the gradient from the next layer (marked as \raisebox{.5pt}{\textcircled{\raisebox{-.9pt} {2}}}
 in Fig. \ref{fig:method:backward}(b)), and the second term is from the first layer of the sharing block at the next timestep (marked as \raisebox{.5pt}{\textcircled{\raisebox{-.9pt} {3}}}
 in Fig. \ref{fig:method:backward}(b)).
The last layer of the sharing block obtains the gradients from the next timestep (marked as \raisebox{.5pt}{\textcircled{\raisebox{-.9pt} {3}}}) which is then, propagated through the intermediate layers.
This allows the weight parameters to be
trained with temporal information, achieving similar performance as the standard SNN architecture.

\noindent\textbf{Gradient of Cross-channel Sharing.}
Assume that we divide the channel into $N$ groups. We define an index set $G = \{1, 2, ..., N\}$.
Then, the gradients of weight parameters in layer $l$ can be computed as:
\begin{equation}
    \begin{aligned}
          \frac{\partial L}{\partial W_l} &= \sum_{t}
      \sum_{i \in G} {\frac{\partial L}{\partial O_l^{t,(i)}}\frac{\partial O_l^{t,(i)}}{\partial U_l^{t,(i)}} \frac{\partial U_l^{t,(i)}}{\partial X^{t}_l}
      \frac{\partial X^{t}_l}{\partial W_l}} \\ 
      & +
      \sum_{t}\sum_{i\in G \backslash \{N\}} \frac{\partial L}{\partial U_l^{t, (i+1)}}  \frac{\partial U_l^{t, (i+1)}}{\partial U_l^{t,(i)}}
      \frac{\partial U_l^{t,(i)}}{\partial X^{t}_l}
      \frac{\partial X^{t}_l}{\partial W_l}
      \\
       & +
       \sum_{t}{
       \frac{\partial L}{\partial U_l^{t+1, (1)}}  \frac{\partial U_l^{t+1, (1)}}{\partial U_l^{t,(N)}}
       \frac{\partial U_l^{t,(N)}}{\partial X^{t}_l}
      \frac{\partial X^{t}_l}{\partial W_l}.
       }
    \end{aligned}
\label{eq:bw_crosschannel}
\end{equation}
The first term represents the gradient from the next layer (marked as \raisebox{.5pt}{\textcircled{\raisebox{-.9pt} {1}}} in Fig. \ref{fig:method:backward}(c)).
The second term is the gradients from the next group's membrane potential except for the last group (marked as \raisebox{.5pt}{\textcircled{\raisebox{-.9pt} {2}}} in Fig. \ref{fig:method:backward}(c)).
The last term represents the gradients from the first group of the next timestep 
(marked as \raisebox{.5pt}{\textcircled{\raisebox{-.9pt} {3}}} in Fig. \ref{fig:method:backward}(c)).
Thus, the gradients propagate through both temporal and spatial dimension, training weight parameters to consider the temporal information.

\begin{figure}[t]
\begin{center}
\def\arraystretch{0.5}
\begin{tabular}{@{\hskip 0.01\linewidth}c@{\hskip 0.01\linewidth}c@{\hskip 0.00\linewidth}c@{\hskip 0.00\linewidth}c@{}c}
 \hspace{-3mm}
\includegraphics[width=0.32\linewidth]{ 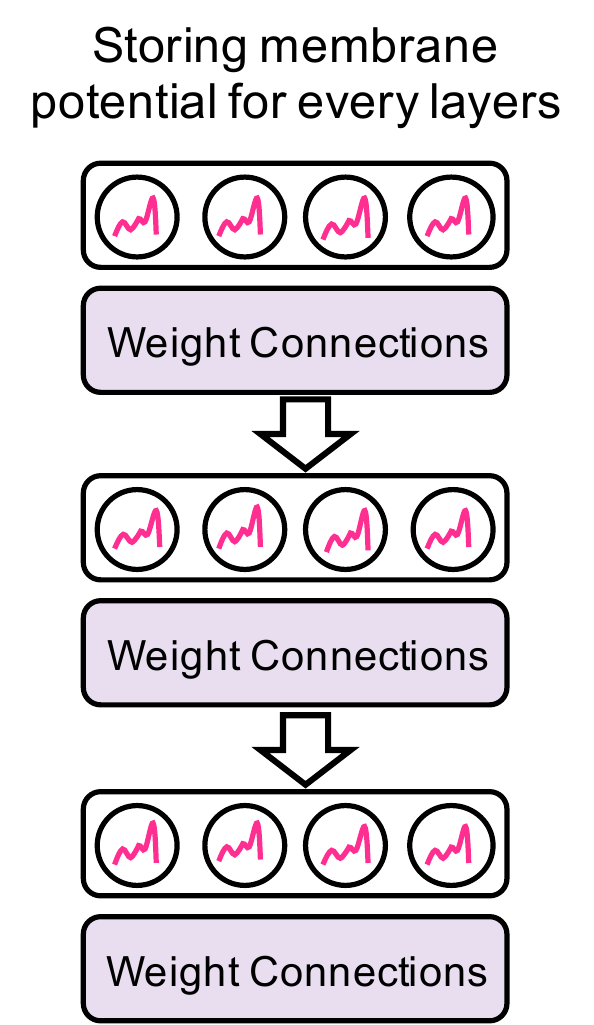} &
\includegraphics[width=0.365\linewidth]{ 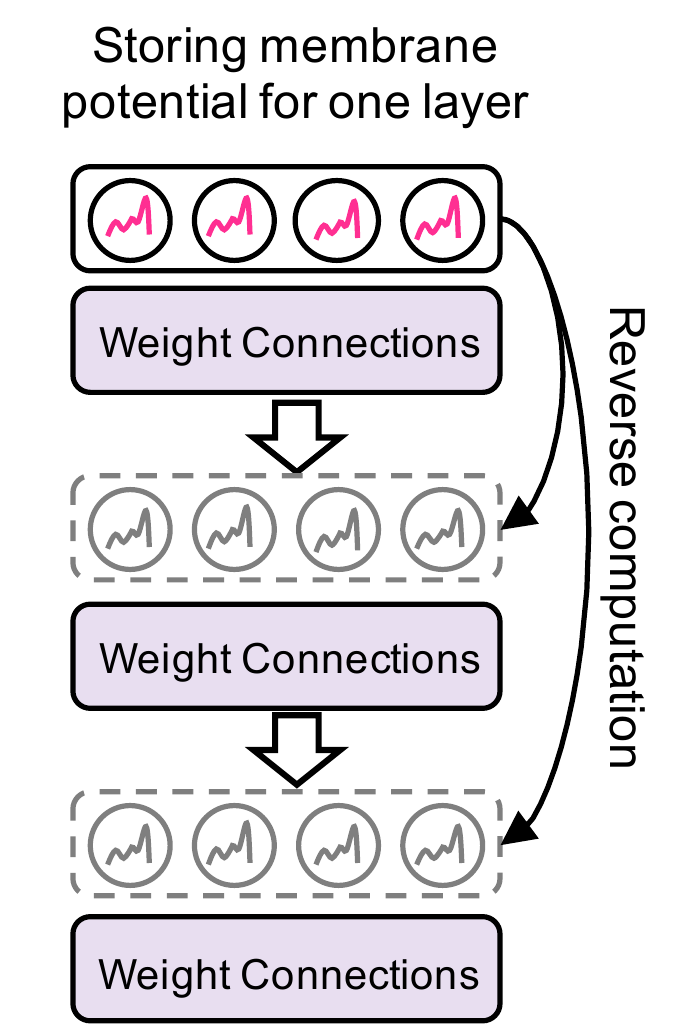} &
\includegraphics[width=0.35\linewidth]{ 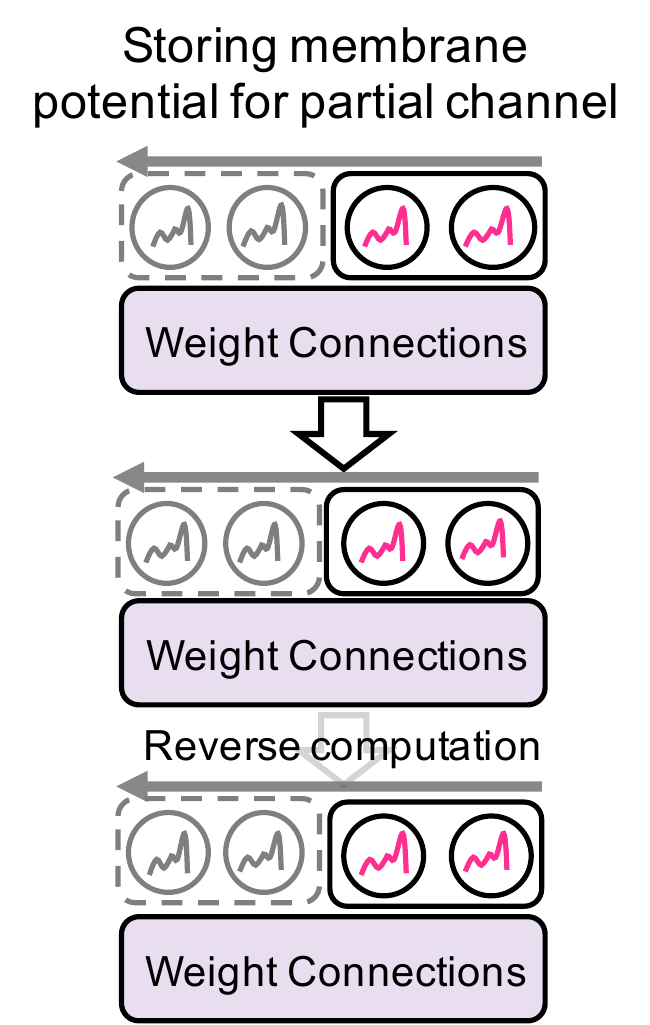} 
\\
\\
 \hspace{-3mm}{(a) Baseline} & \hspace{-4.3mm}
{(b) Cross-layer } & \hspace{-1mm}{(c) Cross-channel}  \\
\end{tabular}
\vspace{-1mm}
\caption{Memory-efficient backpropagation. Compared to baseline, we do not need to store an intermediate membrane potential for backpropagation.
Instead, we perform a reverse computation on the membrane potential from the next layers/channels.
}
 \vspace{-1mm}
\label{fig:method:backward_efficientcomputation}
\end{center}
\end{figure}

\subsection{Memory-Efficient Backpropagation}

In addition to the memory efficiency in forward propagation, our EfficientLIF-Net saves memory costs during backward gradient computation.
As shown in Fig. \ref{fig:method:backward_efficientcomputation}(a), the standard SNNs need to store all membrane potential to compute the gradient such as $\frac{\partial U_l^{t+1}}{\partial U_l^{t}}$ in Eq. \ref{eq:delta_W}.
However, saving the full-precision membrane potential of LIF neurons is costly.

\noindent\textbf{Backpropagation in Cross-layer Sharing.} The key idea here is that the membrane potential of the previous layer can be computed from the next layer's membrane potential in a reverse way (Fig. \ref{fig:method:backward_efficientcomputation}(b)).
Thus, without storing the membrane potential of the intermediate layers during forward, we can compute the backward gradient.
By reorganizing Eq. \ref{eq:LIF_crosslayer} and  Eq. \ref{eq:LIF_crosslayer_last}, we obtain the membrane potential of the previous layer or the previous timestep.
\begin{equation}
\hspace{-2mm}
\begin{cases}
    &  U_{l}^{t}  = \frac{1}{\lambda}(U_{l+1}^{t}-W_{l+1}O_{l}^t) +O_{l}^t. \hspace{10.8mm}\textup{from Eq. \ref{eq:LIF_crosslayer}} \\
    &   U_{l+m}^{t} = \frac{1}{\lambda}(U_{l}^{t+1}-W_{l}O_{l-1}^{t+1})+O_{l+m}^t. \hspace{2mm} \textup{from Eq. \ref{eq:LIF_crosslayer_last}}
    \end{cases}
    \label{eq:LIF_crosslayer_reverse_mem}
\end{equation}

Based on this, we can compute $\frac{\partial U_l^{t+1}}{\partial U_l^{t}}$ in Eq. 
\ref{eq:bw_crosslayer_1}, and $\frac{\partial U_{l}^{t+1}}{\partial U_{l+m}^{t}}$ in Eq. 
\ref{eq:bw_crosslayer_2}, without storing the intermediate membrane potential.

\begin{table*}[h!]
\addtolength{\tabcolsep}{0.5pt}
\centering
\caption{Accuracy and LIF memory cost (Forward \& Backward) comparison between baseline (\ie standard SNN) and our EfficientLIF-Net. Here,
\textit{EfficientLIF-Net[L], EfficientLIF-Net[C\#2], EfficientLIF-Net[L+C\#2]} denote EfficientLIF with cross-layer, cross-channel ($\#$group=2), and cross-layer\&channel sharing, respectively.}
\resizebox{0.65\textwidth}{!}{%
\begin{tabular}{c|lccc}
\toprule
&  \multicolumn{4}{c}{VGG16}     \\
\midrule
\multirow{2}{*}{Dataset} & \multirow{2}{*}{Methods} &\: \:\:\multirow{2}{*}{Acc ($\%$)}\: \:\: & LIF Forward&  LIF Backward   \\
     &  &   &Memory (MB)  &Memory (MB) \\
\midrule 
\midrule 
\multirow{4}{*}{CIFAR10} &  Baseline &91.31 &  1.80 & 9.0  \\
 & EfficientLIF-Net [L] & 90.23& 1.23 & 1.23 \\ 
 & EfficientLIF-Net [C\#2]  &90.30 & 0.90 & 0.90\\ 
 & EfficientLIF-Net [L+C\#2] &90.09 & 0.62 & 0.62  \\
\midrule
\multirow{4}{*}{CIFAR100} &  Baseline& 66.83 &   1.80 & 9.0 \\
 & EfficientLIF-Net [L] &65.01 & 1.23 & 1.23  \\ 
 & EfficientLIF-Net [C\#2] & 64.92 & 0.90 & 0.90  \\ 
 & EfficientLIF-Net [L+C\#2]  & 64.85 & 0.62 & 0.62  \\
\midrule
\multirow{4}{*}{TinyImageNet} &  Baseline& 56.11 &  7.22 & 36.1\\
 & EfficientLIF-Net [L]  & 55.14 & 4.91 & 4.91\\ 
 & EfficientLIF-Net [C\#2] & 55.43 & 3.61 & 3.61  \\ 
 & EfficientLIF-Net [L+C\#2] & 55.36& 2.46 & 2.46  \\
\midrule
\multirow{4}{*}{ImageNet-100} &  Baseline  & 73.81  & 88.43 & 442.15 \\
 & EfficientLIF-Net [L] &  73.22 & 60.10 & 60.10 \\ 
 & EfficientLIF-Net [C\#2] & 72.65&44.21& 44.21 \\ 
 & EfficientLIF-Net [L+C\#2] &72.14& 30.05& 30.05  \\
 \midrule
\multirow{4}{*}{N-Caltech101} &  Baseline & 64.40 & 4.06 & 40.6 \\
 & EfficientLIF-Net [L] & 63.50 &2.76 & 2.76 \\ 
 & EfficientLIF-Net [C\#2] &  64.02  &2.03 & 2.03 \\ 
 & EfficientLIF-Net [L+C\#2] &  63.10 & 1.38& 1.38\\
\bottomrule
\end{tabular}%
}

\vspace{5mm}
\resizebox{0.65\textwidth}{!}{%
\begin{tabular}{c|lccc}
\toprule  &   \multicolumn{4}{c}{ResNet19}   \\
\midrule
\multirow{2}{*}{Dataset} & \multirow{2}{*}{Methods} &\: \:\:\multirow{2}{*}{Acc ($\%$)}\: \:\: & LIF Forward&  LIF Backward \\
     &  &   &Memory (MB)  &Memory (MB)   \\
\midrule 
\midrule 
\multirow{4}{*}{CIFAR10} & Baseline & 92.26   &  2.88  &  14.40 \\
 &  EfficientLIF-Net [L] & 91.99 & 1.31  & 1.31\\ 
 &  EfficientLIF-Net [C\#2] & 91.92 & 1.44 &  1.44 \\ 
 &  EfficientLIF-Net [L+C\#2] & 91.73 & 0.66  & 0.66 \\
\midrule
\multirow{4}{*}{CIFAR100} & Baseline & 70.89 &   2.88  &  14.40 \\
 &EfficientLIF-Net [L] & 70.14 & 1.31 & 1.31\\ 
 & EfficientLIF-Net [C\#2] & 70.01 & 1.44 & 1.44 \\ 
 & EfficientLIF-Net [L+C\#2] & 69.99 & 0.66 & 0.66  \\
\midrule
\multirow{4}{*}{TinyImageNet} &  Baseline & 56.74 &  11.5 & 57.5\\
 &  EfficientLIF-Net [L] & 55.20& 5.25& 5.25\\ 
 &  EfficientLIF-Net [C\#2] &55.44 & 5.75& 5.75 \\ 
 & EfficientLIF-Net [L+C\#2] & 55.10& 2.63& 2.63\\
\midrule
\multirow{4}{*}{ImageNet-100} &   Baseline & 79.38  &140.88 & 704.4\\
 & EfficientLIF-Net [L] &  79.44 & 64.31 & 64.31\\ 
 & EfficientLIF-Net [C\#2] & 78.92 &70.44& 70.44\\ 
 &  EfficientLIF-Net [L+C\#2] &78.88 & 32.16 &32.16  \\
 \midrule
\multirow{4}{*}{N-Caltech101} &  Baseline & 66.27 & 6.47 & 64.7\\
 & EfficientLIF-Net [L] &65.82 & 2.95& 2.95\\ 
 &  EfficientLIF-Net [C\#2] &66.01 & 3.24& 3.24 \\ 
 &  EfficientLIF-Net [L+C\#2] & 65.45 &1.48& 1.48 \\
\bottomrule
\end{tabular}%
}
\vspace{-3mm}
\label{table:exp:accuracy_memory_main}
\end{table*}

\noindent\textbf{Backpropagation in Cross-channel Sharing.}
In a similar way, we can also reduce memory cost through channel dimension by performing a reverse computation on the membrane potential of channel groups (Fig. \ref{fig:method:backward_efficientcomputation}(c)). 
Instead of storing a memory for all channels, we use a partial memory for storing the membrane potential of the last group channel of each layer.
From Eq. \ref{eq:LIF_crosschannel} and Eq. \ref{eq:LIF_crosschannel_nexttime}, we calculate the membrane potential of the previous channel group or the previous timestep.
\begin{equation}
\hspace{-2mm}
\begin{cases}
    &  U_l^{t, (i)} = \frac{1}{\lambda}(  U_l^{t,(i+1)}- X_{l}^{t, (i+1)}) +O_{l}^{t, (i)}. \hspace{4.7mm}\textup{from Eq. \ref{eq:LIF_crosschannel}} \\
    &    U_l^{t, (N)} = \frac{1}{\lambda}(  U_l^{t+1,(1)}-X_{l}^{t+1, (1)}) +O_{l}^{t, (N)}. \hspace{1mm}\textup{from Eq. \ref{eq:LIF_crosschannel_nexttime}}
        \label{eq:LIF_crosschannel_reverse_mem}
    \end{cases}
\end{equation}
This reverse computation allows us to compute $\frac{\partial U_l^{t,(i+1)}}{\partial U_l^{t, (i)}}$ and $\frac{\partial U_l^{t+1,(1)}}{\partial U_l^{t, (N)}}$ in Eq. 
\ref{eq:bw_crosschannel}, without storing the intermediate membrane potential.

\begin{figure}[t]
\begin{center}
\def\arraystretch{0.5}
\begin{tabular}{@{\hskip 0.00\linewidth}c@{\hskip 0.01\linewidth}c@{\hskip 0.00\linewidth}c@{\hskip 0.00\linewidth}c@{}c}
\hspace{-2mm}
\includegraphics[width=0.99\linewidth]{ 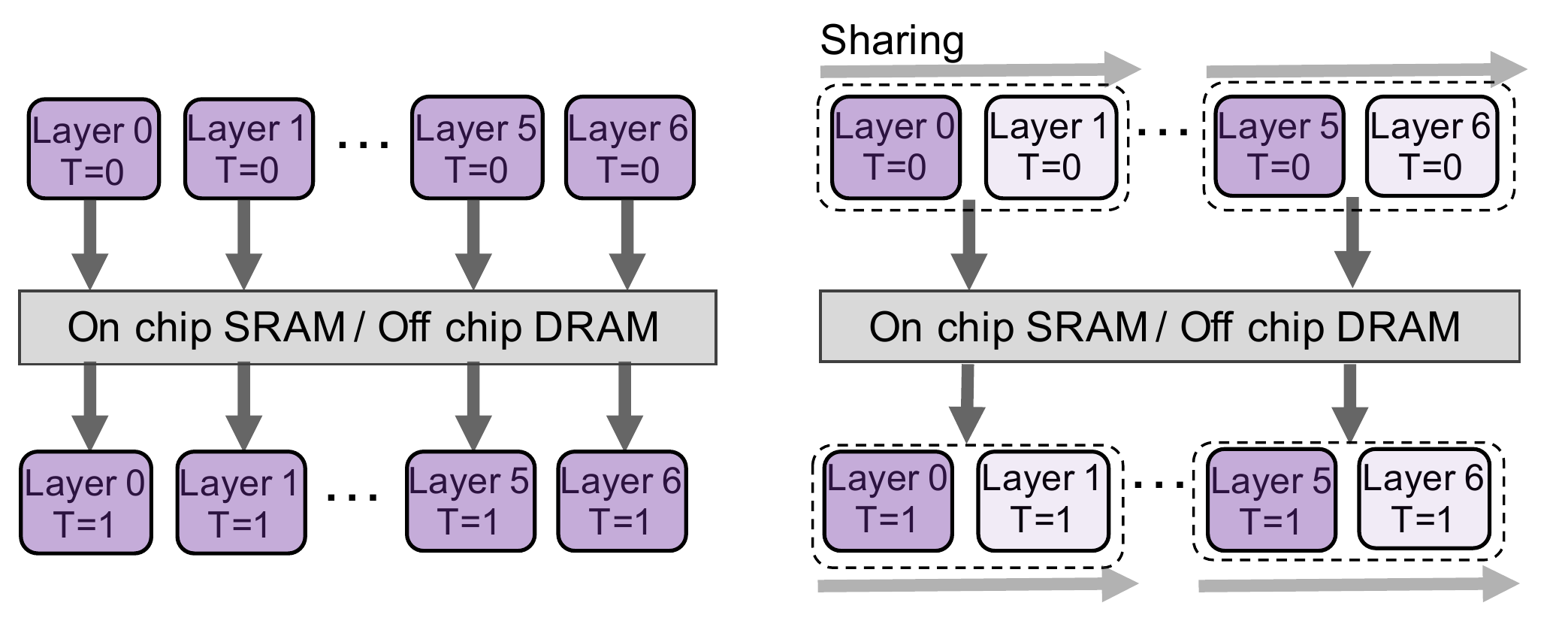} 
\\
 \hspace{2mm}{(a) Cross-layer Sharing} 
 \\
\hspace{-9mm}\includegraphics[width=1.01\linewidth]{ 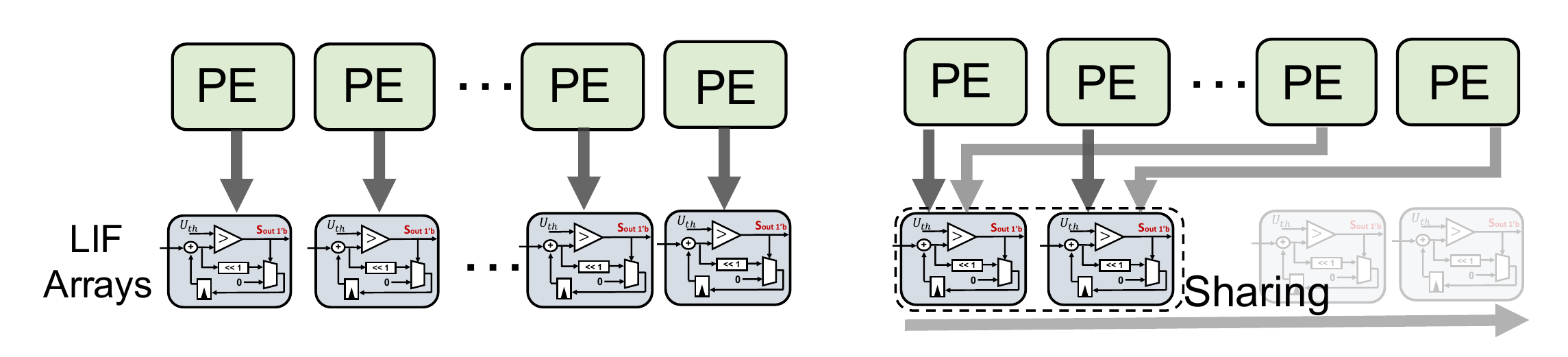} 
\\
 {(b) Cross-channel Sharing } &  \\
\end{tabular}
\caption{Visualization of the potential hardware mapping of the two sharing methods.  We provide some hardware insights on the potential hardware benefits we can get from the EfficientLIF-Net.
}
\label{fig:hardware:insights}
\end{center}
\end{figure}

\subsection{Hardware Disccusion}
\label{hardware_discussion}

In this section, we aim to provide insights into the role that efficientLIF-Net will play during the hardware deployment. 

\noindent\textbf{Cross-layer Sharing.} One of the major contributions that the cross-layer sharing EfficientLIF-Net can make to the hardware is the reduction of memory communications. When deploying an SNN on the hardware, one can either choose to either first process through all the layers and then repeat for all timesteps (standard) or first process through all timesteps then proceed to the next layer (tick-batch~\cite{narayanan2020spinalflow}). While the tick-batch can help to reduce the number of memory communications across timesteps, it requires more hardware resources. On the other hand, with a proper processing pipeline across layers, the standard way of processing SNNs will have smaller hardware resource requirements and a larger throughput. And cross-layer sharing can further reduce the memory communication overheads of the standard SNN processing.

As we show in Fig.~\ref{fig:hardware:insights}(a), instead of writing the membrane potential to the memory for every layer and timestep, layer-sharing EfficientLIF-Net requires only one time of writing to memory for each shared layer for each timestep.

\noindent\textbf{Cross-channel Sharing.} Due to the high level of parallelism and data reuse in these designs, we are focusing on examining the effects of cross-channel sharing on EfficientLIF-Net for ASIC systolic array-based inference accelerators for SNNs~\cite{yin2022sata,narayanan2020spinalflow,lee2022parallel}.

The key idea behind this group of designs is to broadcast input spikes and weights to an array of processing elements (PEs), where accumulators perform convolution operations. Each post-synaptic neuron's entire convolution operation is mapped to one dedicated PE. Once the convolution results are ready, they are sent to the LIF units inside the PE to generate the output spikes.

LIF units are notorious for their high hardware overheads. This is because we need at least one buffer to hold the full precision membrane potential for each neuron. These buffers heavily contribute to the hardware cost of LIF units. Originally, all the prior designs~\cite{yin2022sata,lee2022parallel,narayanan2020spinalflow} equipped each of the 128 PEs with an LIF unit inside to match the design's throughput requirements. These LIF units contribute significantly to the entire PE arrays. Even if the number of LIF units is reduced, there is no way to reduce the number of buffers required to hold the unique membrane potentials for each LIF neuron. 

Based on this design problem, we can instantly realize one role that cross-channel sharing EfficientLIF plays in these hardware platforms. Depending on the number of cross-channel shared LIF neurons, we can have the same ratio of LIF units and buffer reduction at the hardware level, as we show in Fig.~\ref{fig:hardware:insights}(b). For example, in the case of C\#4 shared networks, we can manage to reduce the 128 LIF units in~\cite{yin2022sata,lee2022parallel,narayanan2020spinalflow} to 32. However, the shared LIF units will bring longer latency as a trade-off. In the case of C\#4, originally, one cycle was needed to generate spikes from 128 post-synaptic neurons for one timestep. Now, we will need 4 cycles instead. However, the major portion of latency still lies in the convolution and memory operations, which is typically hundreds of times larger than the cycles needed for generating spikes through LIF units. We provide experimental results in Section~V.C to further illustrate the effects of EfficientLIF-Net on hardware.

\section{Experiments}

\subsection{Implementation Details}

We evaluate our method on four static image datasets (\ie CIFAR10 \cite{krizhevsky2009learning}, CIFAR100 \cite{krizhevsky2009learning}, TinyImageNet \cite{deng2009imagenet}, ImageNet-100 \cite{deng2009imagenet}), and one spiking dataset (\ie N-Caltech101\cite{orchard2015converting}). Here, ImageNet-100 is the subset of ImageNet-1000 dataset\cite{deng2009imagenet}. We use VGG16 \cite{simonyan2014very} and ResNet19  \cite{he2016deep}. 
For both architectures, we use the scaled-up channel size following previous SNN works \cite{zheng2020going,li2022neuromorphic}. 
We train the SNNs with 128 batch samples using SGD optimizer with momentum 0.9 and weight decay 5e-4. 
The initial learning rate is set to 0.1 and decayed with cosine learning rate scheduling~\cite{loshchilov2016sgdr}. 
We set the total number of epochs to 300 for CIFAR10, CIFAR100, and N-Caltech101, and 140 for TinyImageNet and ImageNet-100, respectively. 
We use timesteps $T=5$ across all experiments.

\subsection{Performance Comparison}

Across all experiment sections, 
\textit{EfficientLIF-Net[L]} denotes the  cross-layer sharing scheme,  \textit{EfficientLIF-Net[C\#N]} stands for the cross-channel sharing scheme with $N$ channel groups.
\textit{EfficientLIF-Net[L+C\#N]} means the cross-layer \& channel sharing method.
In Table \ref{table:exp:accuracy_memory_main}, we show the memory benefit from EfficientLIF-Net. We assume a 32-bit representation for membrane potential in LIF neurons. 
{Regarding the backward LIF memory of baseline, we consider the standard backpropagation method which stores membrane potential across entire timesteps \cite{liang2021h2learn,yin2022sata,singh2022skipper}.
%

The experimental results show the following observations:
(1) The EfficientLIF-Net based on ResNet19 achieves a similar performance compared to the baseline, which implies that the proposed membrane sharing strategy still can learn temporal information in spikes.
(2) The EfficientLIF-Net also can be applied to the DVS dataset. 
(3) {
The ResNet19 EfficientLIF-Net achieves less performance degradation compared to VGG16, which implies that skip connection improves training capability in EfficientLIF-Net.
Furthermore, ResNet19 brings higher memory efficiency since it has more layers with similar sized activation.}
(4) As expected, a large-resolution image dataset has more benefits compared to a small-resolution image dataset.
For instance, \textit{EfficientLIF-Net [L+C\#2]} saves $108.72$MB and $672.24$MB for forward and backward path, respectively, on ImageNet-100 which consists of $224\times224$ resolution images, on the other hand, the same architecture saves $2.22$MB (forward) and $13.74$MB (backward) on CIFAR10. 

\subsection{Experimental Analysis}

\noindent \textbf{Analysis on Training Dynamics.}
In our method section, we showed that the backward gradients of each method are different.
To further analyze this, we investigate whether the trained weight parameters can be compatible with other architectures.
We expect that the transferred weights to different architectures may show performance degradation since each architecture has different training dynamics (\eg gradient path).
To this end, we train standard ResNet19-SNN (\ie baseline), EfficientLIF-Net [L], EfficientLIF-Net [C\#2], EfficientLIF-Net [L+C\#2],
In Fig. \ref{fig:exp:transferred_weight}, we report the accuracy of various weights-architecture configurations on CIFAR10 and TinyImageNet.
We observe the following points:
(1) As we expected, transferring weights to a different architecture brings performance degradation. This supports our statement that each architecture has different training dynamics.
(2) Especially, baseline shows a huge performance drop as compared to other architectures. Thus,  EfficientLIF-Net needs to be trained from scratch with gradient-based training.
(3) The trained weights from EfficientLIF-Net [L+C\#2] show reasonable performance on EfficientLIF-Net [L] and EfficientLIF-Net [C] as it contains the feature from both cross-layer and cross-channel sharing.

\begin{figure}[t]
\begin{center}
\def\arraystretch{0.5}
\begin{tabular}{@{\hskip 0.00\linewidth}c@{\hskip 0.01\linewidth}c@{\hskip 0.00\linewidth}c@{\hskip 0.00\linewidth}c@{}c}
\hspace{-2mm}
\includegraphics[width=0.45\linewidth]{ 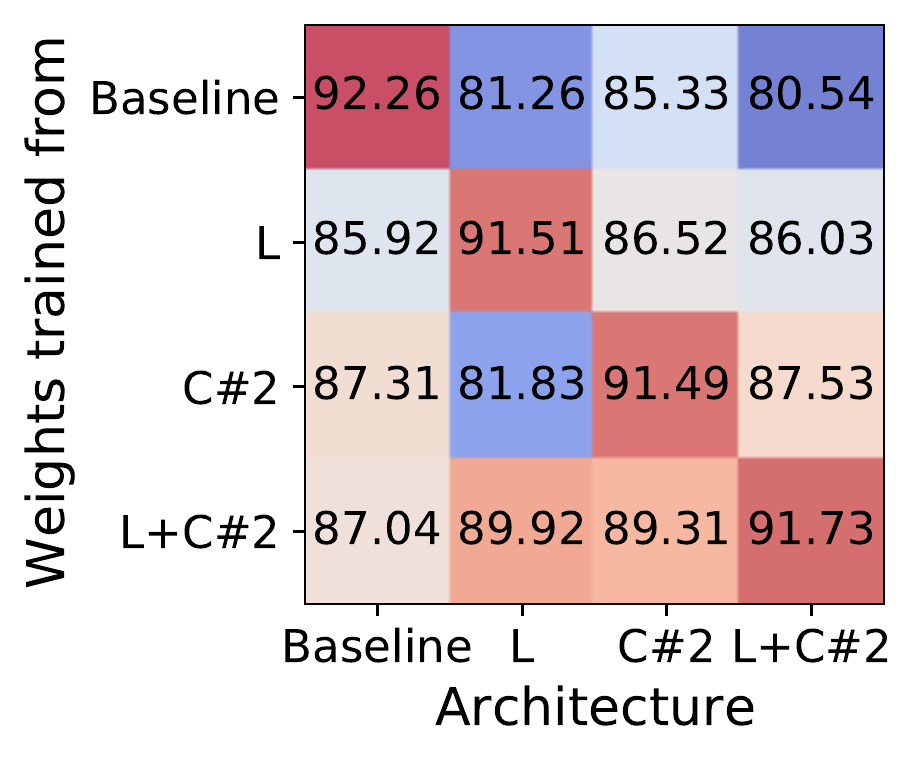} &
\includegraphics[width=0.45\linewidth]{ 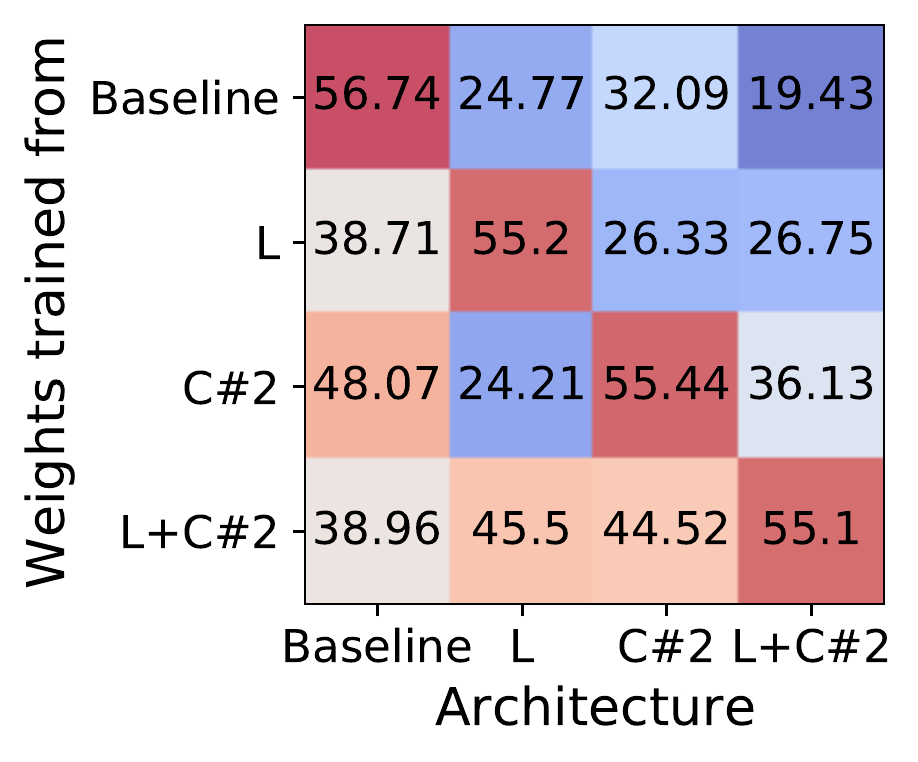} 
\\
 \hspace{9mm}{(a) CIFAR10} &  \hspace{9mm}
{(b) TinyImageNet } &  \\
\end{tabular}
\caption{Analysis on Training Dynamics. Unit: accuracy (\%). We investigate whether
the trained weight parameters can be compatible with other
architectures.
}
\label{fig:exp:transferred_weight}
\end{center}
\end{figure}

\noindent \textbf{Ablation Studies on \#Group.}
In the cross-layer sharing scheme, we can further reduce LIF memory cost by increasing \textit{\#group}.
Table \ref{table:exp:channelablation} shows the accuracy and LIF memory cost with respect to \textit{\#group}.
Interestingly, EfficientLIF-Net with high \textit{\#group} almost maintains the performance while minimizing the LIF memory cost significantly. 
For example, on the ImageNet-100 dataset, EfficientLIF-Net [C\#8] incurs only $0.8\%$ accuracy drop with $75\%$ higher memory saving.
Thus, one can further reduce LIF memory cost by increasing \textit{\#group} based on the hardware requirements.

\begin{table}[t]
\addtolength{\tabcolsep}{0.5pt}
\centering
\caption{Ablation on the number of groups in cross-channel EfficientLIF-Net with ResNet19 architecture.
}
\resizebox{0.48\textwidth}{!}{%
\begin{tabular}{c|ccc}
\toprule
\multirow{2}{*}{Dataset} & \multirow{2}{*}{Methods} & \multirow{2}{*}{Acc ($\%$)} & LIF Memory for \\
     &  &  & Fw \& Bw (MB)  \\
\midrule 
\midrule 
\multirow{3}{*}{CIFAR10}   & EfficientLIF-Net [C\#2] & 91.92  & 1.44 \\ 
 & EfficientLIF-Net [C\#4] & 91.73 & 0.72  \\ 
 & EfficientLIF-Net [C\#8] & 91.21 &  0.36 \\
\midrule
\multirow{3}{*}{TinyImageNet}   & EfficientLIF-Net [C\#2] & 55.44 &  5.75\\ 
 & EfficientLIF-Net [C\#4] & 55.06 & 2.88 \\ 
 & EfficientLIF-Net [C\#8] & 54.84 &  1.44 \\
\midrule
\multirow{3}{*}{ImageNet-100}   & EfficientLIF-Net [C\#2] & 78.92 &  70.44\\ 
 & EfficientLIF-Net [C\#4] & 78.24 &  35.22 \\ 
 & EfficientLIF-Net [C\#8] & 78.12 & 17.61  \\
\bottomrule
\end{tabular}%
}
\label{table:exp:channelablation}
\end{table}

\begin{table}[t]
\addtolength{\tabcolsep}{0.5pt}
\centering
\caption{Performance of combing cross-layer sharing and group convolution on ResNet 19 architecture.
}
\resizebox{0.48\textwidth}{!}{%
\begin{tabular}{c|cccc}
\toprule
{Dataset} & {Methods} & \#Conv. Group &{Acc ($\%$)}  \\

\midrule 
\midrule 
\multirow{3}{*}{CIFAR10}   & EfficientLIF-Net [C\#2]&2 & 91.42   \\ 
 & EfficientLIF-Net [C\#4]&4 & 90.45  \\ 
 & EfficientLIF-Net [C\#8]&8 & 87.38 \\
 \midrule
\multirow{3}{*}{CIFAR100}   & EfficientLIF-Net [C\#2]&2 & 69.26\\ 
 & EfficientLIF-Net [C\#4]&4 & 66.42 \\ 
 & EfficientLIF-Net [C\#8]&8 & 60.20 \\
\midrule
\multirow{3}{*}{TinyImageNet}   & EfficientLIF-Net [C\#2]&2 & 53.65 \\ 
 & EfficientLIF-Net [C\#4]&4 &  51.39 \\ 
 & EfficientLIF-Net [C\#8]&8 & 42.86 \\
\bottomrule
\end{tabular}%
}
\label{table:exp:channelgroupconvablation}
\end{table}

\noindent \textbf{Combining with Group Convolution.}
To further enhance the efficiency in cross-channel sharing, we explore the feasibility of combining a group convolution layer with cross-layer sharing.
Since group convolution splits input channels and output channels into multiple groups, they can be applied to each channel spike ($O_l^{t,(i)}$ in Eq. \ref{eq:LIF_crosschannel_outconcat}).
In Table \ref{table:exp:channelgroupconvablation}, we observe the accuracy does not show a huge drop with two convolution groups.
However, as the number of groups increases, the performance goes down drastically due to lesser number of parameters available for training convergence.

\noindent \textbf{Soft Reset vs. Hard Reset.}
We also conduct experiments on the reset scheme in our EfficientLIF-Net.
The membrane potential can be reset to zero (\ie hard reset), or decreased by the threshold value (\ie soft reset). 
In Table \ref{table:exp:resetablation}, we compare the accuracy of both reset schemes on ResNet19 architecture, where we observe the hard reset achieves similar accuracy as the soft reset. 
However, using the hard reset does not allow computation of the previous layer's or timestep's membrane potential in a reverse way (Eq. \ref{eq:LIF_crosslayer_reverse_mem} and Eq. \ref{eq:LIF_crosschannel_reverse_mem}) during backpropagation. This is because the hard reset removes the residual membrane potential which can be used in the reverse computation.
Therefore, our EfficientLIF-Net is based on the soft reset such that we get memory savings both during forward and backward.


\noindent \textbf{Analysis on Spike Rate.}
In Fig. \ref{fig:exp_spikerate}, we compare the spike rate across all different LIF sharing schemes in ResNet19.
We conduct experiments on four datasets.
Note, a high spike rate implies the networks require larger computational cost.
The experimental results show that all LIF sharing schemes have a similar spike rate as the baseline. 
This demonstrates that EfficientLIF-Net does not bring further computational overhead while saving memory cost by sharing the membrane potential.

\noindent \textbf{Time Overhead Analysis.}
We measured the time overhead on a V100 GPU with a batch size of 128. We used VGG16 with CIFAR10 and ImageNet-100 datasets with image sizes of 32x32 and 224x224, respectively. Tabel \ref{table:exp:timeover} shows the latency results for each method.
Interestingly, we found that our method improves computation time, implying that our LIF layer-sharing method reduces the time required to access DRAM, which originally takes a significant percentage of computational time. As a result, our method can be implemented without a huge computational burden.

\begin{table}[t]
\addtolength{\tabcolsep}{0.5pt}
\centering
\caption{Analysis on the computational time.
}
\vspace{-2mm}
\resizebox{0.4\textwidth}{!}{%
\begin{tabular}{l|cc}
\toprule
Method (latency: ms)	& 32x32 & 224x224  \\
\midrule 
\midrule 
Baseline& 105.12 & 148.21 \\
   EfficientLIF-Net [L]& 79.21 & 131.25   \\
  EfficientLIF-Net [C\#2]& 80.62 & 142.75  \\ 
  EfficientLIF-Net [L+C\#2]& 81.26 & 143.05 \\
\bottomrule
\end{tabular}%
}
\label{table:exp:timeover}
\end{table}

\noindent \textbf{Memory Cost Breakdown.}
In Fig. \ref{fig:memorycostanalysis}, we compare the memory cost breakdown between the SNN baseline and EfficientLIF-Net in both forward and backward.
In the memory cost comparison, we consider memory for weight parameters (32-bit), spike activation (1-bit), and LIF neurons (32-bit). 
In the baseline SNN, LIF neurons take a dominant portion for both forward and backward memory cost. Especially, for backward, LIF neurons occupy around $7\times$ larger memory than weights or activation memory.
Our EfficientLIF-Net significantly reduces the LIF memory cost, resulting in less memory overhead compared to weight parameters (in both foward and backward) and activation (in backward only).

\begin{table}[t]
\addtolength{\tabcolsep}{0.5pt}
\centering
\caption{Ablation on the reset methods.
}
\vspace{-2mm}
\resizebox{0.48\textwidth}{!}{%
\begin{tabular}{c|lccc}
\toprule
{Dataset} & {Methods} & Reset Scheme &{Acc ($\%$)}  \\

\midrule 
\midrule 
\multirow{6}{*}{CIFAR10}   & 
   EfficientLIF-Net [L]& Soft & 91.99   \\
 & EfficientLIF-Net [L]& Hard & 91.66  \\ 
 & EfficientLIF-Net [C\#2]& Soft & 91.92  \\ 
  & EfficientLIF-Net [C\#2]& Hard & 91.67  \\ 
 & EfficientLIF-Net [L+C\#2]& Soft & 91.73 \\
  & EfficientLIF-Net [L+C\#2]& Hard & 91.65 \\
 \midrule
\multirow{6}{*}{CIFAR100}   & 
   EfficientLIF-Net [L]& Soft & 70.14   \\
 & EfficientLIF-Net [L]& Hard & 70.05   \\ 
 & EfficientLIF-Net [C\#2]& Soft & 70.01  \\ 
  & EfficientLIF-Net [C\#2]& Hard & 68.93 \\ 
 & EfficientLIF-Net [L+C\#2]& Soft & 69.99 \\
  & EfficientLIF-Net [L+C\#2]& Hard & 69.74 \\
\bottomrule
\end{tabular}%
}
\label{table:exp:resetablation}
\end{table}

\begin{figure}[h]
\begin{center}
\def\arraystretch{0.5}
\begin{tabular}{@{}c@{\hskip 0.03\linewidth}c@{}c}
\includegraphics[width=0.90\linewidth]{ 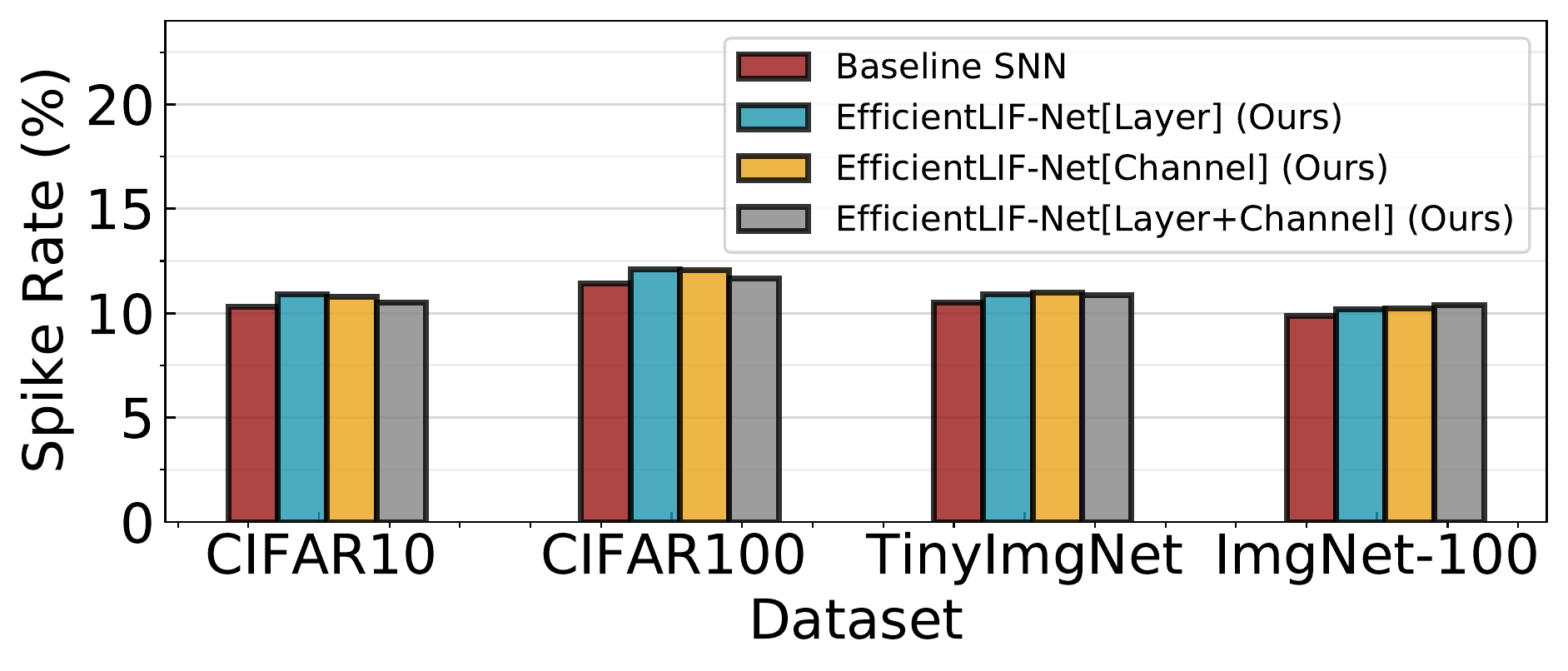} 
 \vspace{-5mm}
\end{tabular}
\end{center}
\caption{
Spike rate analysis on four public datasets.}
\label{fig:exp_spikerate}
\end{figure}

\noindent \textbf{EfficientLIF-Net with Weight Pruning.}
As pruning for SNNs is popular due to its usage on edge devices \cite{chen2021pruning,neftci2016stochastic,guo2020unsupervised,shi2019soft,kim2022exploring}, it is important to figure out whether the advantage from EfficientLIF-Net remains in sparse SNNs.
Before exploring the effectiveness of the LIF sharing method in sparse SNNs, we first investigate if LIF neurons still require a huge memory in sparse SNNs. This is because a number of LIF neurons might not generate output spikes in the high weight sparsity regime ($\ge 90\%$), then, the memory cost for such dead neurons can be reduced.
To this end, we prune the SNN model to varied sparsity using magnitude-based pruning \cite{han2015learning}.
Interestingly, as shown in Fig. \ref{fig:exp:pruning} \textbf{Left}, only $\sim 3 \%$ neurons do not generate spikes (\ie dead neuron) across all sparsity levels.
This implies that the LIF memory cost is still problematic in sparse SNNs. 
Based on the observation, we prune EfficientLIF-Net and compare the memory cost and accuracy with the standard SNN baseline. Here, we prune all architectures to have $94.94\%$ weight sparsity. 
In Fig. \ref{fig:exp:pruning} \textbf{Right}, the baseline architecture requires 2.9 MB for LIF neurons, which is equivalent to $\sim60\%$ of the memory cost for weight parameters.
With cross-layer (denoted as \textit{L} in Fig. \ref{fig:exp:pruning})  and cross-channel sharing (denoted as \textit{C\#2} in Fig. \ref{fig:exp:pruning}), we can reduce the LIF memory cost by about half compared to the baseline.
Cross-layer\& channel sharing (denoted as \textit{L+C\#2} in Fig. \ref{fig:exp:pruning}) further reduces the memory cost, which takes only $\sim 23\%$ memory compared to the baseline.
Overall, the results demonstrate that LIF memory reduction is not only important for high-resolution images but also for relatively low-resolution images such as CIFAR10 especially when considering pruned SNNs.

\begin{figure}[t]
\begin{center}
\def\arraystretch{0.5}
\begin{tabular}{@{}c@{\hskip 0.03\linewidth}c@{}c}

\includegraphics[width=0.9\linewidth]{ 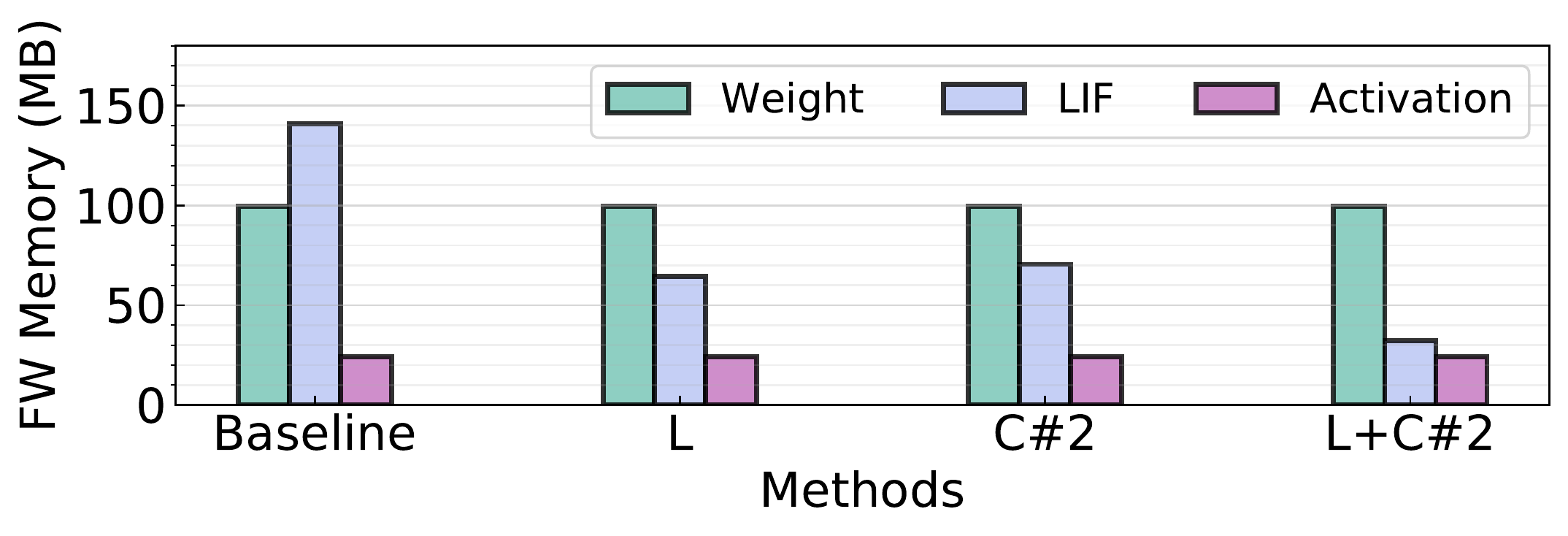}  \vspace{-1mm} \\
\includegraphics[width=0.9\linewidth]{ 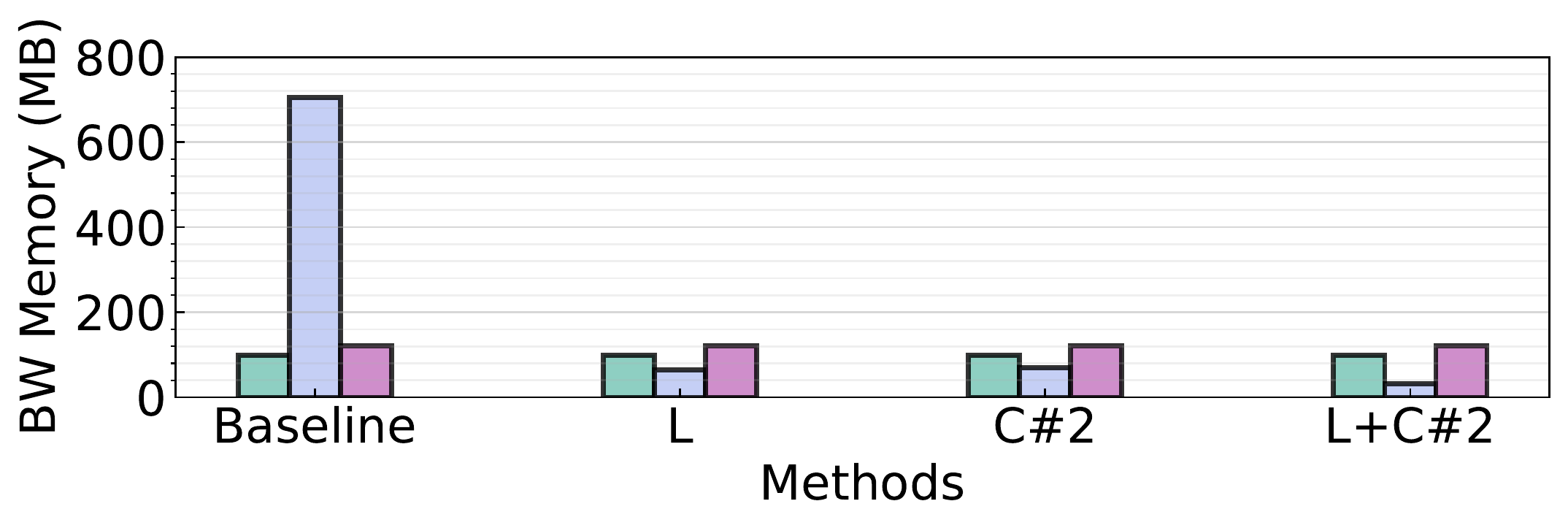} 
 \vspace{-5mm}
\end{tabular}
\end{center}
\caption{
Comparison of the memory breakdown between the baseline SNN and the EfficientLIF-Net in both forward and backward. We use ResNet19 architecture on ImageNet-100.}
 \vspace{-1mm}
\label{fig:memorycostanalysis}
\end{figure}

\begin{figure}[t]
\begin{center}
\def\arraystretch{0.5}
\begin{tabular}{@{\hskip 0.00\linewidth}c@{\hskip 0.01\linewidth}c@{\hskip 0.00\linewidth}c@{\hskip 0.00\linewidth}c@{}c}
\hspace{-3mm}
\includegraphics[width=0.44\linewidth]{ 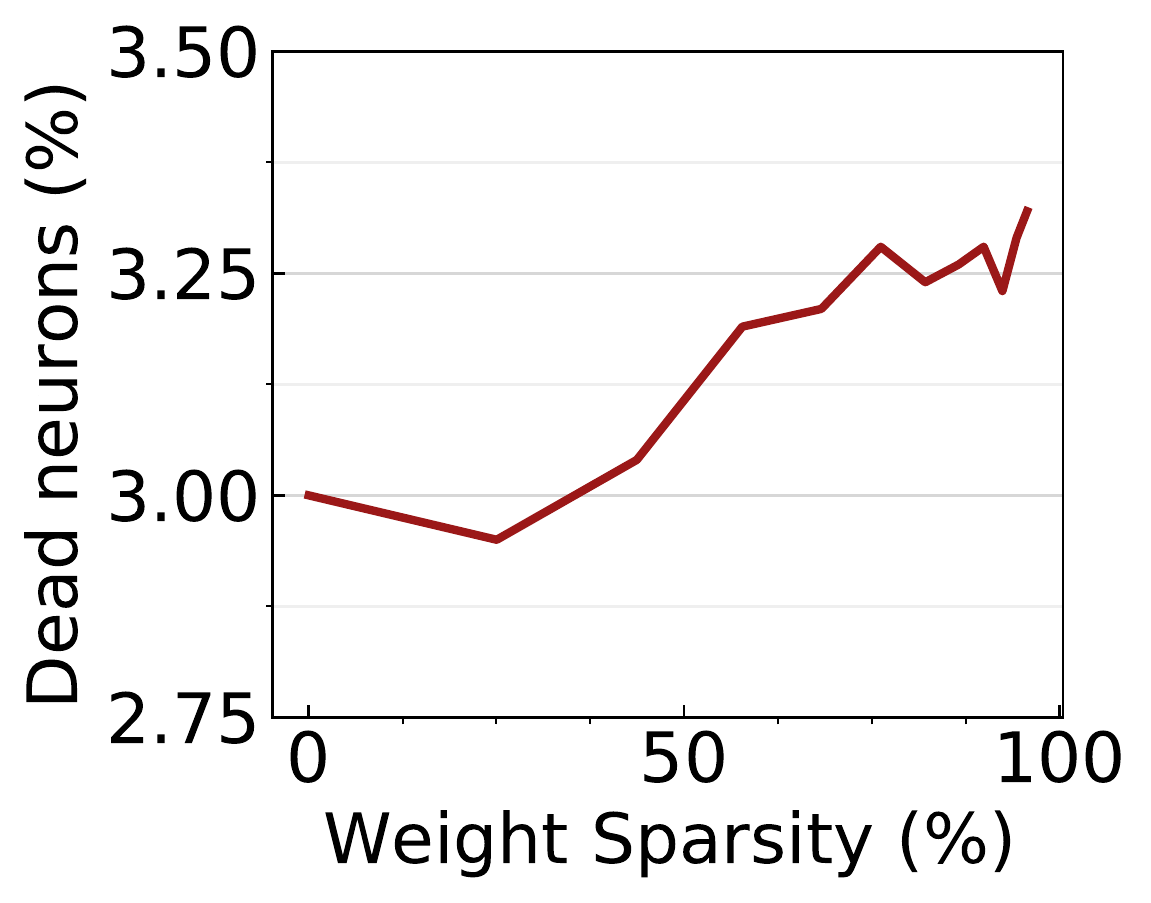} &
\includegraphics[width=0.505\linewidth]{ 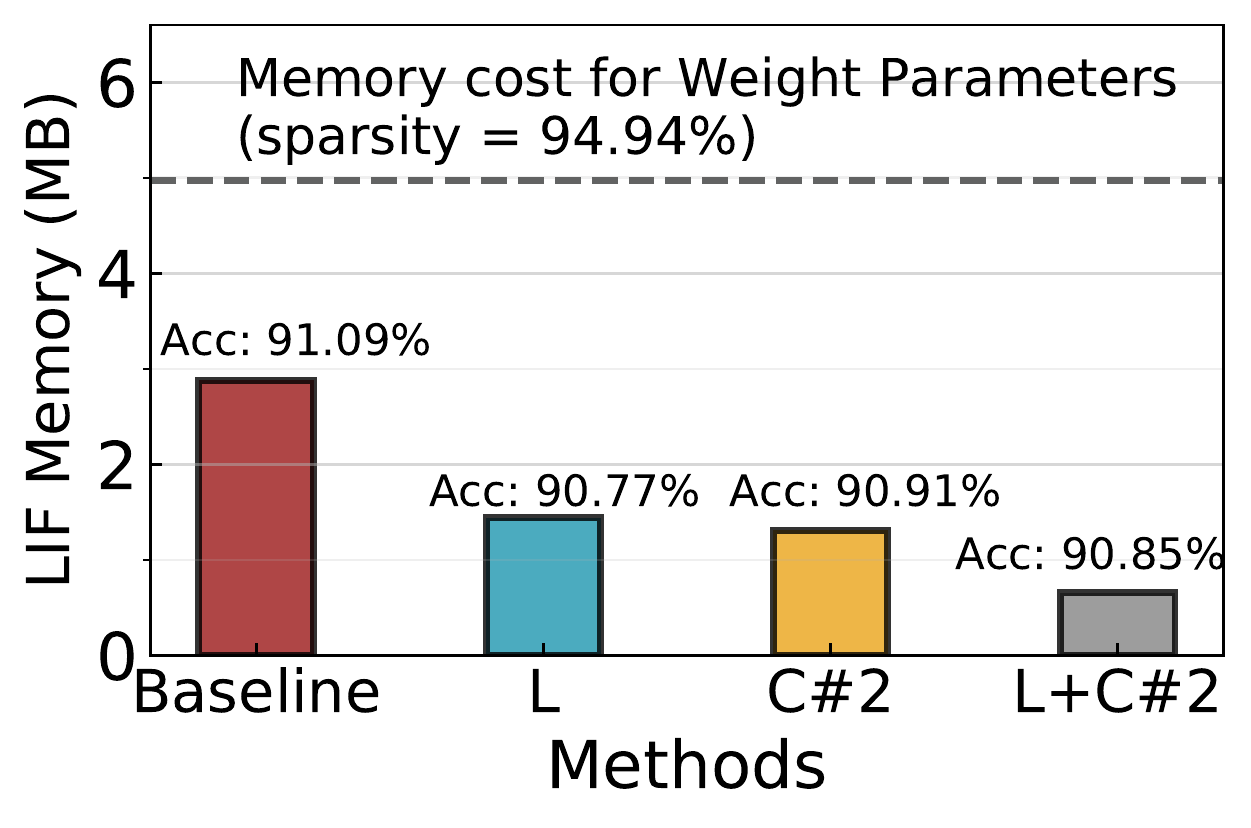} 
\end{tabular}
\caption{Experiments on ResNet19 EfficientLIF-Net with weight pruning methods on CIFAR10. \textbf{Left:} Most LIF neurons generate output spikes although the weight sparsity increases. Therefore, the LIF memory cost cannot be reduced by weight pruning. \textbf{Right:} Accuracy and LIF memory cost comparison across baseline and EfficientLIF-Net. The weight memory cost across all models is $\sim 5 MB$ indicated with a grey dotted line.
}
\label{fig:exp:pruning}
\end{center}
\end{figure}

\begin{figure}[t]
\begin{center}
\def\arraystretch{0.5}
\begin{tabular}{@{}c@{\hskip 0.03\linewidth}c@{}c}

\includegraphics[width=0.9\linewidth]{ 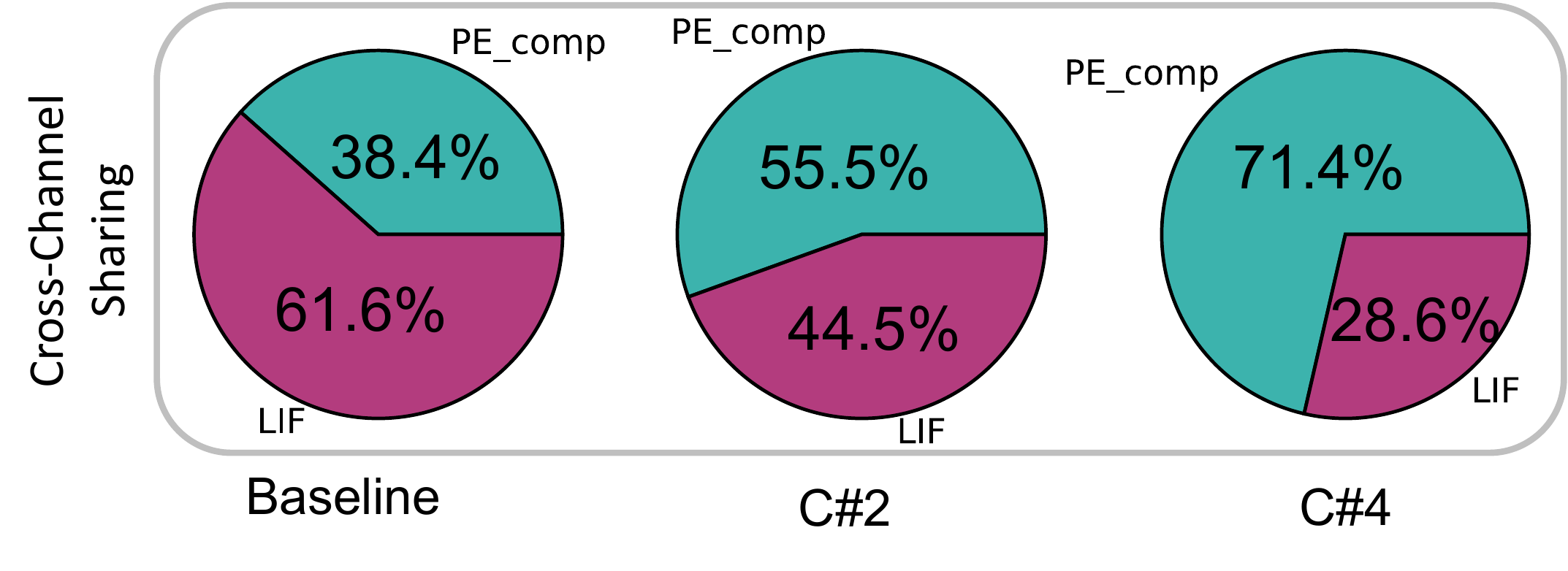}  \vspace{-1mm} \\
\includegraphics[width=0.9\linewidth]{ 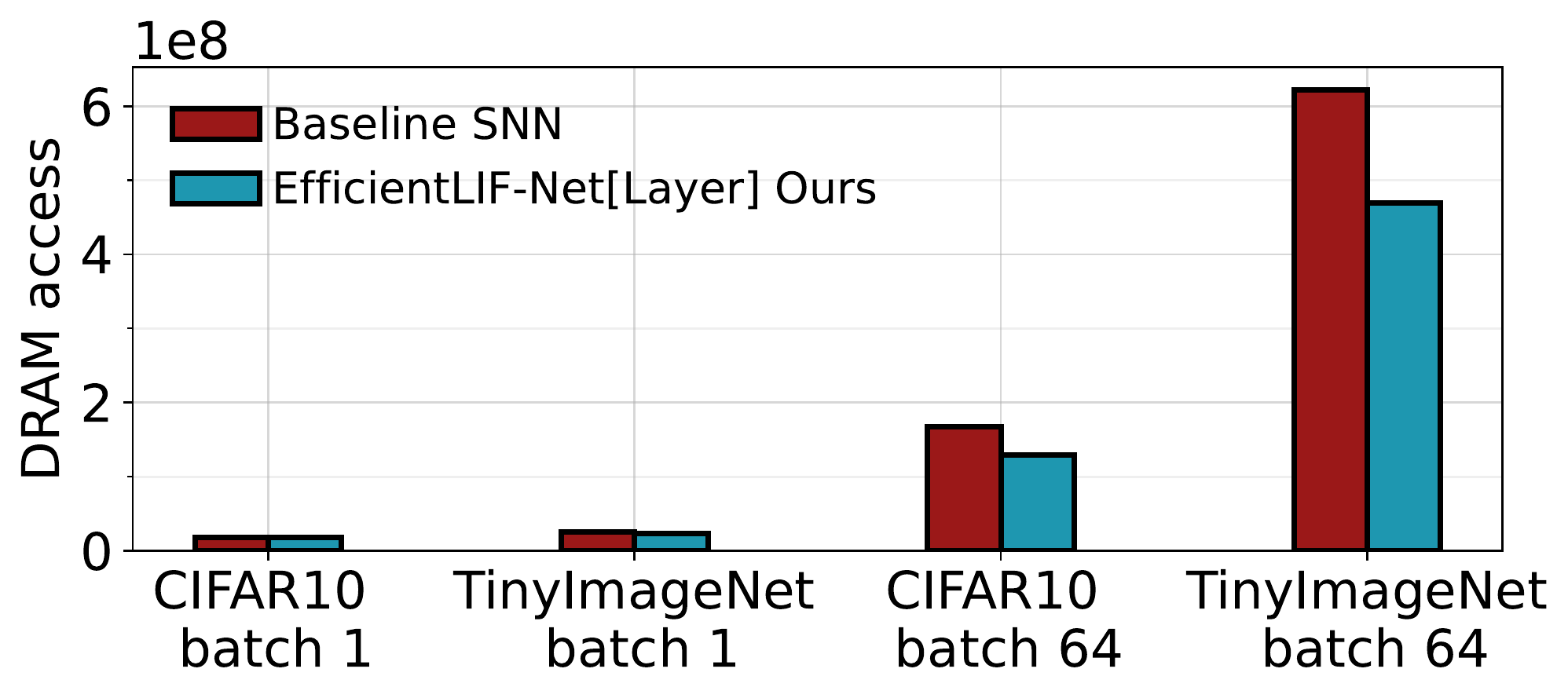} 
 \vspace{-3mm}
\end{tabular} 
\end{center}
\caption{
\textbf{Top}: The area breakdown of computation units for the Baseline SNN, EfficientLIF-Net[C\#2], and EfficientLIF-Net[C\#4] in a 128 PE system.
\textbf{Bottom}: Comparison of DRAM access reduction between the Baseline SNN and EfficientLIF-Net[Layer] on VGG-16 across various datasets. The reduction is contrasted for single batch processing and multiple mini-batch processing scenarios.}
 \vspace{-1mm}
\label{fig:hardware_evaluation}
\end{figure}

\noindent \textbf{Hardware Evaluation.}
\label{exp_hardware}
As discussed in Section \ref{hardware_discussion}, both cross-channel and cross-layer sharing can significantly enhance hardware efficiency during deployment. From the top portion of Fig.\ref{fig:hardware_evaluation}, it is evident that cross-channel sharing in EfficientLIF-Net can considerably decrease the number of required LIF units. Specifically, our approach reduces the area portion of LIF units from 61.6\% to 28.6\% of total computation units when employing C\#4 cross-channel sharing.

The bottom part of Fig.\ref{fig:hardware_evaluation} indicates that cross-layer sharing can effectively minimize the number of DRAM accesses, which is the most energy-consuming operation during on-chip SNN inference. For single-batch scenarios, the reduction is not significant, since weight data movement dominates the DRAM accesses, as outlined in \cite{yin2022sata}. However, when employing mini-batches, the reduction becomes more substantial. We note a 23\% and 25\% reduction in total DRAM accesses on CIFAR10 and TinyImageNet, respectively, for 64 mini-batches. This reduction trend continues to rise with larger mini-batch numbers.

\subsection{Evaluation on Human Activity Recognition (HAR) datasets}

To further validate our method on datasets that rely heavily on temporal information, we conduct experiments using Human Activity Recognition (HAR) datasets obtained from wearable devices. Descriptions of these datasets are provided below:
\begin{itemize}
\item \textit{UCI-HAR} \cite{anguita2013public} consists of 10.3k instances collected from 30 subjects, involving six different activities: walking, walking upstairs, walking downstairs, sitting, standing, and lying. The dataset employs sensors such as a 3-axis accelerometer and a 3-axis gyroscope (both at 50Hz) from a Samsung Galaxy SII.
\item \textit{HHAR} \cite{stisen2015smart} is collected from nine subjects and encompasses six daily activities: biking, sitting, standing, walking, stair ascent, and stair descent. The dataset utilizes accelerometers from eight smartphones and four smartwatches (with sampling rates ranging from 50Hz to 200Hz).
\end{itemize}
Following previous work, we split both datasets into 64\% for the training set, 16\% for the validation set, and 20\% for the test set. We report test accuracy when the model achieves its best validation accuracy.

In Table \ref{table:exp:har}, we compare our method with the baseline model, which consists of six 1D-convolutional layers, \ie $Conv1D(InputChannel, 128) - 4 \times Conv1D(128,128) – Conv1D(128, \#Class)$. In addition, we provide the performance of other methods  \cite{aviles2019coarse,mukherjee2020ensemconvnet,wang2020human} on HHAR and UCI-HAR. Aviles \etal \cite{aviles2019coarse} uses a CNN, Mukherjee \etal \cite{mukherjee2020ensemconvnet} uses a combination of CNN and LSTM, and Wang \etal \cite{wang2020human} uses an LSTM. 
From the table, we can observe the following results:
(1) The baseline Spiking MLP achieves an accuracy of 97.68\% on the HHAR dataset and 96.06\% on the UCI-HAR dataset, which is comparable accuracy with the previous  methods.
(2)Comparing the different configurations of EfficientLIF-Net to the baseline Spiking MLP, we can see that the EfficientLIF-Net maintains a similar level of accuracy as the baseline on both datasets. 
These results suggest that our LIF-sharing method also works well with tasks that heavily rely on temporal information.
Overall, our empirical results support the observation that gradients propagate through both temporal and spatial dimensions, effectively training the weight parameters to account for temporal information, as demonstrated in Eq. \ref{eq:bw_crosslayer_1}, \ref{eq:bw_crosslayer_2}, and \ref{eq:bw_crosschannel}.

\begin{table}[t]
\addtolength{\tabcolsep}{0.5pt}
\centering
\caption{Accuracy (\%) comparison between baseline (\ie 6 layer 1D-Convolutional SNN) and our EfficientLIF-Net.
Here, EfficientLIF-Net[L], EfficientLIF-Net[C\#2], EfficientLIF-Net[L+C\#2] denote EfficientLIF with cross-layer, cross-channel (\#group=2),
and cross-layer\&channel sharing, respectively.
}
\resizebox{0.43\textwidth}{!}{%
\begin{tabular}{l|cc}
\toprule
Method / Dataset	& HHAR \cite{stisen2015smart}  & UCI-HAR  \cite{anguita2013public} \\
\midrule 
\midrule 
Aviles \etal \cite{aviles2019coarse} & 96.19 &		96.29   \\
Mukherjee \etal \cite{mukherjee2020ensemconvnet} & 	97.15 &		97.87   \\
Wang \etal \cite{wang2020human} & 95.59 &		82.41  \\
\midrule 
Spiking MLP (Baseline) & 97.68 &	96.06 \\
   EfficientLIF-Net [L]& 97.10 &	95.58   \\
  EfficientLIF-Net [C\#2]& 97.68 &	96.06  \\ 
  EfficientLIF-Net [L+C\#2]& 97.10 & 95.04  \\
\bottomrule
\end{tabular}%
}
\label{table:exp:har}
\end{table}

\section{Conclusion}

In this paper, we highlight and tackle the problem of LIF memory cost in SNNs. This problem becomes severe as the image resolution increases.
To address this, we propose EfficientLIF-Net where we share the membrane potential across layers and channels, which can effectively reduce memory usage.
During backpropagation, our EfficientLIF-Net also enables reverse computation on the previous layer and channel.
Therefore, we only need to store the membrane potential of the last layer/channel during forward.
In our experiments, EfficientLIF-Net achieves similar performance and computational cost  while significantly reducing memory cost compared to standard SNN baseline.
We also found that the LIF memory problem exists in sparse-weight SNNs where even a small resolution dataset causes LIF memory overhead.
The memory benefit of EfficientLIF-Net is shown in pruned SNNs, which implies our method is complementary to previous compression methods.

\section*{Acknowledgements}
This work was supported in part by CoCoSys, a JUMP2.0 center sponsored by DARPA and SRC, Google Research Scholar Award, the National Science Foundation CAREER Award, TII (Abu Dhabi), the DARPA AI Exploration (AIE) program, and the DoE MMICC center SEA-CROGS (Award \#DE-SC0023198).



\bibliographystyle{elsarticle-num} 
\bibliography{egbib}

\begin{thebibliography}{10}
\expandafter\ifx\csname url\endcsname\relax
  \def\url#1{\texttt{#1}}\fi
\expandafter\ifx\csname urlprefix\endcsname\relax\def\urlprefix{URL }\fi
\expandafter\ifx\csname href\endcsname\relax
  \def\href#1#2{#2} \def\path#1{#1}\fi

\bibitem{roy2019towards}
K.~Roy, A.~Jaiswal, P.~Panda, Towards spike-based machine intelligence with
  neuromorphic computing, Nature 575~(7784) (2019) 607--617.

\bibitem{christensen20222022}
D.~V. Christensen, R.~Dittmann, B.~Linares-Barranco, A.~Sebastian, M.~Le~Gallo,
  A.~Redaelli, S.~Slesazeck, T.~Mikolajick, S.~Spiga, S.~Menzel, et~al., 2022
  roadmap on neuromorphic computing and engineering, Neuromorphic Computing and
  Engineering (2022).

\bibitem{wu2018spatio}
Y.~Wu, L.~Deng, G.~Li, J.~Zhu, L.~Shi, Spatio-temporal backpropagation for
  training high-performance spiking neural networks, Frontiers in neuroscience
  12 (2018) 331.

\bibitem{wu2019direct}
Y.~Wu, L.~Deng, G.~Li, J.~Zhu, Y.~Xie, L.~Shi, Direct training for spiking
  neural networks: Faster, larger, better, in: Proceedings of the AAAI
  Conference on Artificial Intelligence, Vol.~33, 2019, pp. 1311--1318.

\bibitem{kundu2021hire}
S.~Kundu, M.~Pedram, P.~A. Beerel, Hire-snn: Harnessing the inherent robustness
  of energy-efficient deep spiking neural networks by training with crafted
  input noise, in: Proceedings of the IEEE/CVF International Conference on
  Computer Vision, 2021, pp. 5209--5218.

\bibitem{fang2021deep}
W.~Fang, Z.~Yu, Y.~Chen, T.~Huang, T.~Masquelier, Y.~Tian, Deep residual
  learning in spiking neural networks, arXiv preprint arXiv:2102.04159 (2021).

\bibitem{liu2001spike}
Y.-H. Liu, X.-J. Wang, Spike-frequency adaptation of a generalized leaky
  integrate-and-fire model neuron, Journal of computational neuroscience 10~(1)
  (2001) 25--45.

\bibitem{akopyan2015truenorth}
F.~Akopyan, J.~Sawada, A.~Cassidy, R.~Alvarez-Icaza, J.~Arthur, P.~Merolla,
  N.~Imam, Y.~Nakamura, P.~Datta, G.-J. Nam, et~al., Truenorth: Design and tool
  flow of a 65 mw 1 million neuron programmable neurosynaptic chip, IEEE
  transactions on computer-aided design of integrated circuits and systems
  34~(10) (2015) 1537--1557.

\bibitem{davies2018loihi}
M.~Davies, N.~Srinivasa, T.-H. Lin, G.~Chinya, Y.~Cao, S.~H. Choday, G.~Dimou,
  P.~Joshi, N.~Imam, S.~Jain, et~al., Loihi: A neuromorphic manycore processor
  with on-chip learning, IEEE Micro 38~(1) (2018) 82--99.

\bibitem{furber2014spinnaker}
S.~B. Furber, F.~Galluppi, S.~Temple, L.~A. Plana, The spinnaker project,
  Proceedings of the IEEE 102~(5) (2014) 652--665.

\bibitem{orchard2021efficient}
G.~Orchard, E.~P. Frady, D.~B.~D. Rubin, S.~Sanborn, S.~B. Shrestha, F.~T.
  Sommer, M.~Davies, Efficient neuromorphic signal processing with loihi 2, in:
  2021 IEEE Workshop on Signal Processing Systems (SiPS), IEEE, 2021, pp.
  254--259.

\bibitem{he2016deep}
K.~He, X.~Zhang, S.~Ren, J.~Sun, Deep residual learning for image recognition,
  in: CVPR, 2016, pp. 770--778.

\bibitem{liang2021h2learn}
L.~Liang, Z.~Qu, Z.~Chen, F.~Tu, Y.~Wu, L.~Deng, G.~Li, P.~Li, Y.~Xie, H2learn:
  High-efficiency learning accelerator for high-accuracy spiking neural
  networks, arXiv preprint arXiv:2107.11746 (2021).

\bibitem{singh2022skipper}
S.~Singh, A.~Sarma, S.~Lu, A.~Sengupta, M.~T. Kandemir, E.~Neftci,
  V.~Narayanan, C.~R. Das, Skipper: Enabling efficient snn training through
  activation-checkpointing and time-skipping, in: 2022 55th IEEE/ACM
  International Symposium on Microarchitecture (MICRO), IEEE, 2022, pp.
  565--581.

\bibitem{yin2022sata}
R.~Yin, A.~Moitra, A.~Bhattacharjee, Y.~Kim, P.~Panda, Sata: Sparsity-aware
  training accelerator for spiking neural networks, arXiv preprint
  arXiv:2204.05422 (2022).

\bibitem{sengupta2019going}
A.~Sengupta, Y.~Ye, R.~Wang, C.~Liu, K.~Roy, Going deeper in spiking neural
  networks: Vgg and residual architectures, Frontiers in neuroscience 13 (2019)
  95.

\bibitem{han2020rmp}
B.~Han, G.~Srinivasan, K.~Roy, Rmp-snn: Residual membrane potential neuron for
  enabling deeper high-accuracy and low-latency spiking neural network, in:
  Proceedings of the IEEE/CVF Conference on Computer Vision and Pattern
  Recognition, 2020, pp. 13558--13567.

\bibitem{diehl2015fast}
P.~U. Diehl, D.~Neil, J.~Binas, M.~Cook, S.-C. Liu, M.~Pfeiffer,
  Fast-classifying, high-accuracy spiking deep networks through weight and
  threshold balancing, in: 2015 International Joint Conference on Neural
  Networks (IJCNN), ieee, 2015, pp. 1--8.

\bibitem{rueckauer2017conversion}
B.~Rueckauer, I.-A. Lungu, Y.~Hu, M.~Pfeiffer, S.-C. Liu, Conversion of
  continuous-valued deep networks to efficient event-driven networks for image
  classification, Frontiers in neuroscience 11 (2017) 682.

\bibitem{li2021free}
Y.~Li, S.~Deng, X.~Dong, R.~Gong, S.~Gu, A free lunch from ann: Towards
  efficient, accurate spiking neural networks calibration, arXiv preprint
  arXiv:2106.06984 (2021).

\bibitem{lee2016training}
J.~H. Lee, T.~Delbruck, M.~Pfeiffer, Training deep spiking neural networks
  using backpropagation, Frontiers in neuroscience 10 (2016) 508.

\bibitem{lee2020enabling}
C.~Lee, S.~S. Sarwar, P.~Panda, G.~Srinivasan, K.~Roy, Enabling spike-based
  backpropagation for training deep neural network architectures, Frontiers in
  Neuroscience 14 (2020).

\bibitem{neftci2019surrogate}
E.~O. Neftci, H.~Mostafa, F.~Zenke, Surrogate gradient learning in spiking
  neural networks, IEEE Signal Processing Magazine 36 (2019) 61--63.

\bibitem{shrestha2018slayer}
S.~B. Shrestha, G.~Orchard, Slayer: Spike layer error reassignment in time,
  arXiv preprint arXiv:1810.08646 (2018).

\bibitem{wu2021training}
H.~Wu, Y.~Zhang, W.~Weng, Y.~Zhang, Z.~Xiong, Z.-J. Zha, X.~Sun, F.~Wu,
  Training spiking neural networks with accumulated spiking flow, ijo 1~(1)
  (2021).

\bibitem{li2021differentiable}
Y.~Li, Y.~Guo, S.~Zhang, S.~Deng, Y.~Hai, S.~Gu, Differentiable spike:
  Rethinking gradient-descent for training spiking neural networks, Advances in
  Neural Information Processing Systems 34 (2021) 23426--23439.

\bibitem{kim2022neural}
Y.~Kim, Y.~Li, H.~Park, Y.~Venkatesha, P.~Panda, Neural architecture search for
  spiking neural networks, arXiv preprint arXiv:2201.10355 (2022).

\bibitem{wu2020progressive}
J.~Wu, C.~Xu, D.~Zhou, H.~Li, K.~C. Tan, Progressive tandem learning for
  pattern recognition with deep spiking neural networks, arXiv preprint
  arXiv:2007.01204 (2020).

\bibitem{kim2021revisiting}
Y.~Kim, P.~Panda, Revisiting batch normalization for training low-latency deep
  spiking neural networks from scratch, Frontiers in neuroscience (2021) 1638.

\bibitem{kim2022exploringb}
Y.~Kim, Y.~Li, H.~Park, Y.~Venkatesha, A.~Hambitzer, P.~Panda, Exploring
  temporal information dynamics in spiking neural networks, arXiv preprint
  arXiv:2211.14406 (2022).

\bibitem{venkatesha2021federated}
Y.~Venkatesha, Y.~Kim, L.~Tassiulas, P.~Panda, Federated learning with spiking
  neural networks, arXiv preprint arXiv:2106.06579 (2021).

\bibitem{yang2022lead}
H.~Yang, K.-Y. Lam, L.~Xiao, Z.~Xiong, H.~Hu, D.~Niyato, H.~Vincent~Poor, Lead
  federated neuromorphic learning for wireless edge artificial intelligence,
  Nature communications 13~(1) (2022) 1--12.

\bibitem{skatchkovsky2020federated}
N.~Skatchkovsky, H.~Jang, O.~Simeone, Federated neuromorphic learning of
  spiking neural networks for low-power edge intelligence, in: ICASSP 2020-2020
  IEEE International Conference on Acoustics, Speech and Signal Processing
  (ICASSP), IEEE, 2020, pp. 8524--8528.

\bibitem{neftci2016stochastic}
E.~O. Neftci, B.~U. Pedroni, S.~Joshi, M.~Al-Shedivat, G.~Cauwenberghs,
  Stochastic synapses enable efficient brain-inspired learning machines,
  Frontiers in neuroscience 10 (2016) 241.

\bibitem{rathi2018stdp}
N.~Rathi, P.~Panda, K.~Roy, Stdp-based pruning of connections and weight
  quantization in spiking neural networks for energy-efficient recognition,
  IEEE Transactions on Computer-Aided Design of Integrated Circuits and Systems
  38~(4) (2018) 668--677.

\bibitem{guo2020unsupervised}
W.~Guo, M.~E. Fouda, H.~E. Yantir, A.~M. Eltawil, K.~N. Salama, Unsupervised
  adaptive weight pruning for energy-efficient neuromorphic systems, Frontiers
  in Neuroscience (2020) 1189.

\bibitem{shi2019soft}
Y.~Shi, L.~Nguyen, S.~Oh, X.~Liu, D.~Kuzum, A soft-pruning method applied
  during training of spiking neural networks for in-memory computing
  applications, Frontiers in neuroscience 13 (2019) 405.

\bibitem{deng2021comprehensive}
L.~Deng, Y.~Wu, Y.~Hu, L.~Liang, G.~Li, X.~Hu, Y.~Ding, P.~Li, Y.~Xie,
  Comprehensive snn compression using admm optimization and activity
  regularization, IEEE transactions on neural networks and learning systems
  (2021).

\bibitem{chen2021pruning}
Y.~Chen, Z.~Yu, W.~Fang, T.~Huang, Y.~Tian, Pruning of deep spiking neural
  networks through gradient rewiring, arXiv preprint arXiv:2105.04916 (2021).

\bibitem{kim2022exploring}
Y.~Kim, Y.~Li, H.~Park, Y.~Venkatesha, R.~Yin, P.~Panda, Exploring lottery
  ticket hypothesis in spiking neural networks, in: Computer Vision--ECCV 2022:
  17th European Conference, Tel Aviv, Israel, October 23--27, 2022,
  Proceedings, Part XII, Springer Nature Switzerland Cham, 2022, pp. 102--120.

\bibitem{li2022quantization}
C.~Li, L.~Ma, S.~B. Furber, Quantization framework for fast spiking neural
  networks, Frontiers in Neuroscience (2022) 1055.

\bibitem{meng2022training}
Q.~Meng, M.~Xiao, S.~Yan, Y.~Wang, Z.~Lin, Z.-Q. Luo, Training high-performance
  low-latency spiking neural networks by differentiation on spike
  representation, in: Proceedings of the IEEE/CVF Conference on Computer Vision
  and Pattern Recognition, 2022, pp. 12444--12453.

\bibitem{guo2022reducing}
Y.~Guo, Y.~Chen, L.~Zhang, Y.~Wang, X.~Liu, X.~Tong, Y.~Ou, X.~Huang, Z.~Ma,
  Reducing information loss for spiking neural networks, in: Computer
  Vision--ECCV 2022: 17th European Conference, Tel Aviv, Israel, October
  23--27, 2022, Proceedings, Part XI, Springer, 2022, pp. 36--52.

\bibitem{datta2022hoyer}
G.~Datta, Z.~Liu, P.~A. Beerel, Hoyer regularizer is all you need for ultra
  low-latency spiking neural networks, arXiv preprint arXiv:2212.10170 (2022).

\bibitem{schaefer2020quantizing}
C.~J. Schaefer, S.~Joshi, Quantizing spiking neural networks with integers, in:
  International Conference on Neuromorphic Systems 2020, 2020, pp. 1--8.

\bibitem{chowdhury2021spatio}
S.~S. Chowdhury, I.~Garg, K.~Roy, Spatio-temporal pruning and quantization for
  low-latency spiking neural networks, in: 2021 International Joint Conference
  on Neural Networks (IJCNN), IEEE, 2021, pp. 1--9.

\bibitem{lui2021hessian}
H.~W. Lui, E.~Neftci, Hessian aware quantization of spiking neural networks,
  in: International Conference on Neuromorphic Systems 2021, 2021, pp. 1--5.

\bibitem{fang2021incorporating}
W.~Fang, Z.~Yu, Y.~Chen, T.~Masquelier, T.~Huang, Y.~Tian, Incorporating
  learnable membrane time constant to enhance learning of spiking neural
  networks, in: Proceedings of the IEEE/CVF International Conference on
  Computer Vision, 2021, pp. 2661--2671.

\bibitem{deng2009imagenet}
J.~Deng, W.~Dong, R.~Socher, L.-J. Li, K.~Li, L.~Fei-Fei, Imagenet: A
  large-scale hierarchical image database, in: 2009 IEEE conference on computer
  vision and pattern recognition, Ieee, 2009, pp. 248--255.

\bibitem{narayanan2020spinalflow}
S.~Narayanan, K.~Taht, R.~Balasubramonian, E.~Giacomin, P.-E. Gaillardon,
  Spinalflow: An architecture and dataflow tailored for spiking neural
  networks, in: 2020 ACM/IEEE 47th Annual International Symposium on Computer
  Architecture (ISCA), IEEE, 2020, pp. 349--362.

\bibitem{lee2022parallel}
J.-J. Lee, W.~Zhang, P.~Li, Parallel time batching: Systolic-array acceleration
  of sparse spiking neural computation, in: 2022 IEEE International Symposium
  on High-Performance Computer Architecture (HPCA), IEEE, 2022, pp. 317--330.

\bibitem{krizhevsky2009learning}
A.~Krizhevsky, G.~Hinton, et~al., Learning multiple layers of features from
  tiny images (2009).

\bibitem{orchard2015converting}
G.~Orchard, A.~Jayawant, G.~K. Cohen, N.~Thakor, Converting static image
  datasets to spiking neuromorphic datasets using saccades, Frontiers in
  neuroscience 9 (2015) 437.

\bibitem{simonyan2014very}
K.~Simonyan, A.~Zisserman, Very deep convolutional networks for large-scale
  image recognition, ICLR (2015).

\bibitem{zheng2020going}
H.~Zheng, Y.~Wu, L.~Deng, Y.~Hu, G.~Li, Going deeper with directly-trained
  larger spiking neural networks, arXiv preprint arXiv:2011.05280 (2020).

\bibitem{li2022neuromorphic}
Y.~Li, Y.~Kim, H.~Park, T.~Geller, P.~Panda, Neuromorphic data augmentation for
  training spiking neural networks, arXiv preprint arXiv:2203.06145 (2022).

\bibitem{loshchilov2016sgdr}
I.~Loshchilov, F.~Hutter, Sgdr: Stochastic gradient descent with warm restarts,
  arXiv preprint arXiv:1608.03983 (2016).

\bibitem{han2015learning}
S.~Han, J.~Pool, J.~Tran, W.~Dally, Learning both weights and connections for
  efficient neural network, Advances in neural information processing systems
  28 (2015).

\bibitem{anguita2013public}
D.~Anguita, A.~Ghio, L.~Oneto, X.~Parra, J.~L. Reyes-Ortiz, et~al., A public
  domain dataset for human activity recognition using smartphones., in: Esann,
  Vol.~3, 2013, p.~3.

\bibitem{stisen2015smart}
A.~Stisen, H.~Blunck, S.~Bhattacharya, T.~S. Prentow, M.~B. Kj{\ae}rgaard,
  A.~Dey, T.~Sonne, M.~M. Jensen, Smart devices are different: Assessing and
  mitigatingmobile sensing heterogeneities for activity recognition, in:
  Proceedings of the 13th ACM conference on embedded networked sensor systems,
  2015, pp. 127--140.

\bibitem{aviles2019coarse}
C.~Avil{\'e}s-Cruz, A.~Ferreyra-Ram{\'\i}rez, A.~Z{\'u}{\~n}iga-L{\'o}pez,
  J.~Villegas-Cort{\'e}z, Coarse-fine convolutional deep-learning strategy for
  human activity recognition, Sensors 19~(7) (2019) 1556.

\bibitem{mukherjee2020ensemconvnet}
D.~Mukherjee, R.~Mondal, P.~K. Singh, R.~Sarkar, D.~Bhattacharjee,
  Ensemconvnet: a deep learning approach for human activity recognition using
  smartphone sensors for healthcare applications, Multimedia Tools and
  Applications 79 (2020) 31663--31690.

\bibitem{wang2020human}
L.~Wang, R.~Liu, Human activity recognition based on wearable sensor using
  hierarchical deep lstm networks, Circuits, Systems, and Signal Processing 39
  (2020) 837--856.

\end{thebibliography}





\end{document}